\documentclass{article}
\usepackage[final,nonatbib]{neurips_2022}
\usepackage[svgnames]{xcolor}
\usepackage[utf8]{inputenc}
\usepackage[T1]{fontenc}
\usepackage{hyperref}
\usepackage{url}
\usepackage{booktabs}
\usepackage{amsfonts}
\usepackage{nicefrac}
\usepackage{microtype}
\usepackage[pdftex]{graphicx}

\usepackage{adjustbox}
\usepackage{amsmath}
\usepackage{mathtools}
\usepackage{bbm}
\usepackage{caption}
\usepackage{makecell}
\usepackage{microtype}
\usepackage{multicol}
\usepackage{multirow}
\usepackage{siunitx}
\usepackage{subcaption}
\usepackage{wrapfig}
\usepackage{xspace}
\usepackage{hyperref}
\usepackage[capitalize]{cleveref}
\usepackage[numbers,square,sort]{natbib}
\usepackage{bm}
\usepackage{algorithm}
\usepackage{algpseudocode}
\usepackage{soul}
\usepackage[normalem]{ulem}
\hypersetup{colorlinks=true, urlcolor=magenta, citecolor=green}

\newcommand{\gray}[1]{\textcolor{darkgray}{#1}}
\newcommand{\movia}{MOVi-A\xspace}
\newcommand{\movic}{MOVi-C\xspace}

\DeclarePairedDelimiterX{\infdivx}[2]{(}{)}{%
  #1\;\delimsize\|\;#2%
}
\newcommand{\kldiv}{D_\text{KL}\infdivx}

\makeatletter
\newcommand{\ie}{\textit{i}.\textit{e}.\@ifnextchar{,}{}{~}}
\makeatother
\makeatletter
\newcommand{\eg}{\textit{e}.\textit{g}.\@ifnextchar{,}{}{~}}
\makeatother

\makeatletter
\renewcommand{\paragraph}{%
  \@startsection{paragraph}{4}%
  {\z@}{0em}{-1em}%
  {\normalfont\normalsize\bfseries}%
}
\makeatother

\definecolor{lightgrey}{rgb}{0.43,0.43,0.43}
\hypersetup{colorlinks=true, urlcolor=magenta, citecolor=green, linkcolor=blue}

\title{Unsupervised Multi-object Segmentation \\ by Predicting Probable Motion Patterns}
\author{Laurynas Karazija\thanks{Authors contributed equally.}$^*$, Subhabrata Choudhury$^*$, \\ 
\textbf{Iro Laina, Christian Rupprecht, Andrea Vedaldi} \vspace{0.5em}\\
Visual Geometry Group\\
University of Oxford\\
Oxford, UK\\
{\tt\small \{laurynas,subha,iro,chrisr,vedaldi\}@robots.ox.ac.uk}
}

\begin{document}
\maketitle
\begin{abstract}
We propose a new approach to learn to segment multiple image objects without manual supervision.
The method can extract objects form still images, but uses videos for supervision.
While prior works have considered motion for segmentation, a key insight is that, while motion can be used to identify objects, not all objects are necessarily in motion: the absence of motion does not imply the absence of objects.
Hence, our model learns to predict image regions that are likely to contain motion patterns characteristic of objects moving rigidly.
It does not predict \emph{specific} motion, which cannot be done unambiguously from a still image, but a distribution of possible motions, which includes the possibility that an object does not move at all.
We demonstrate the advantage of this approach over its deterministic counterpart and show state-of-the-art unsupervised object segmentation performance on simulated and real-world benchmarks, surpassing methods that use motion even at test time. 
As our approach is applicable to variety of network architectures that segment the scenes, we also apply it to existing image reconstruction-based models showing drastic improvement.
Project page and code: \url{https://www.robots.ox.ac.uk/~vgg/research/ppmp}.
\end{abstract}
\section{Introduction}\label{s:introduction}

Humans have an innate ability to segment individual objects in a picture,
but learning this capability with an algorithm usually relies on manual supervision.
In this paper, we consider the problem of learning to segment objects from visual data only\,---\,without externally provided labels.
Algorithms for this task usually assume that objects are seen in different configurations and in front of different backgrounds.
They then exploit cues such as the visual consistency and the co-occurrence of characteristic object parts to learn to discover and segment individual object instances.

Most such methods use still images as input and train with a reconstruction objective~\cite{greff2019iodine,locatello2020object,jiang2020generative}.
They work well on simple synthetic scenes, but they struggle in scenes with more complex visual appearance~\cite{karazija2021clevrtex}.
This has motivated the development of algorithms that use videos as input and can thus observe the motion of the objects as evidence of their presence.
A common way of using motion for unsupervised learning is to seek for a compact representation
to \emph{reconstruct} the video itself~~\cite{veerapaneni2020entity,jiang2020scalor,Lin2020ImprovingGI, Kabra2021SIMONeVT}.
Effectively, such methods seek for a compressed representation of appearance, but do not sidestep entirely the difficult task of modeling it.
This has motivated  authors to look instead at reconstructing the video's \emph{optical flow}~\cite{yang2021self-supervised,kipf2022conditional}.
In fact, the optical flow measures the motion of the objects directly and is much simpler to model than the objects' appearance.

In this paper, we propose a method that lies in-between these two classes, \ie single image-based and video-based.
Our model learns to segment objects in still images, and is thus based on appearance, but learns to do so using \emph{video as a learning signal}, in an unsupervised manner.
The learning process can be summarized as follows.
Given an image, each pixel is assigned to a slot that represents a certain object.
The quality of the assignments is then measured by the coherence of the (unobserved) optical flow within the extracted regions.
Because predicting optical flow from a still image is intrinsically ambiguous, the method models \emph{distributions} of probable flow patterns within each region.
The idea is that (rigid) objects generate characteristic flow patterns that can be used to distinguish between them.

Note that, because the segmentation network is based on a single image, it will learn to partition all objects contained in the scene, not just the ones that actually move in the video, \ie solving an object instance segmentation problem, rather than \emph{motion} segmentation.

We derive closed-form distributions for the flow field generated by rigid objects moving in the scene.
We also derive efficient expressions for the calculation of the flow probability under such models.
The problem of decomposing the image into a number of regions is cast as a standard image segmentation task and an off-the-shelf neural network can be used for it.

As our method uses videos to train an image-based model, we introduce two new datasets which are straightforward video extensions of the existing image datasets CLEVR~\citep{johnson2017clevr} and \textsc{ClevrTex}~\citep{karazija2021clevrtex}.
These datasets are built by animating the objects with initial velocities and using a physics simulation to generate realistic object movement.
Our datasets are constructed with the realistic assumption that not all objects are moving at all times.
This means that motion alone cannot be used as the sole cue for objectness and reflects scenarios such as a workbench where a person only interacts with a small number of objects at a time. 

Empirically, we validate our model against several ablations and baselines. 
We compare our approach to existing unsupervised multi-object segmentation methods achieving state-of-the-art performance.
We demonstrate particularly strong performance in visually complex scenes even with unseen objects and textures at test time.
Our experiments in comparison to \emph{image-based} models and, in particular, adding our motion-aware formulation to existing models shows substantial improvements, confirming that motion is an important cue to learn objectness.
Furthermore, we show that our learned segmenter, which operates on still images, produces better segmentation results than current \emph{video-based} methods that use motion information at test time.
Finally, we also apply our method to real-world self-driving scenarios where we show superior performance to prior work.
\section{Related Work}\label{s:related}
\paragraph{Multi-object decomposition.}
Learning unsupervised object segmentation for static scenes is a well-researched problem in computer vision~\cite{eslami2016attend,lin2020space, crawford2019spatially, jiang2020generative, burgess2019monet, emami2021efficient,engelcke2019genesis, engelcke2021genesis, locatello2020object, smirnov2021marionette, greff2016tagger, Greff2015BindingVR, stelzner19faster, kun2019deep}. These methods aim to decompose the scene into constituent parts, \eg the different foreground objects and the background. 
Glimpse-based methods~\cite{eslami2016attend,lin2020space, crawford2019spatially, jiang2020generative} find input patches (glimpses) that contain the objects in the scene. These methods learn object descriptors that encode their properties  (\eg position, number, size of the objects) using variational inference, composing glimpses into the final picture.
More related to ours are approaches that learn per-pixel object masks~\cite{burgess2019monet, emami2021efficient,engelcke2019genesis, engelcke2021genesis, locatello2020object}.  MoNet~\cite{burgess2019monet} and IODINE~\cite{greff2019iodine} employ multiple encoding-decoding steps to sequentially explain the scene as a collection of regions. Slot Attention~\cite{locatello2020object} uses a multi-step soft clustering-like scheme to find the regions simultaneously. 
In all cases, learning is posed as an image reconstruction problem. 
In order to align learnable slots with semantic objects, models have to make efficient use of a limited representation available for each region, such as learning to only explain visual appearance.
This principle, however, is difficult to extend to visually complex data~\citep{karazija2021clevrtex} and relies on custom specialized architectures.
Instead, our method allows for any standard segmentation architecture to be used, which we train to predict regions that are most likely described by rigid motion patterns.%

\paragraph{Video-based multi object  decomposition.}
Another line of work extends the unsupervised object decomposition problem to videos~\cite{van2018relational, kosiorek2018sequential, He2019TrackingBA, Weis2021, zablotskaia21aPROVIDE, Kabra2021SIMONeVT, Li2020LearningOR, kossen2020structured, Singh2021StructuredWB, Wu2021APEXUO, Lin2020ImprovingGI, jiang2020scalor}. 
Many of these methods work mainly with simpler datasets~\cite{kossen2020structured, zablotskaia21aPROVIDE, Singh2021StructuredWB} 
and require sequential frames for training.
For example, SCALOR~\cite{jiang2020scalor} is a glimpse-based method that discovers and propagates objects across frames to learn intermediate object latents. SIMONe~\cite{Kabra2021SIMONeVT} processes the whole video at once, learning both temporal scene representation and time-invariant object representations simultaneously. 
Slot Attention for Video (SAVi)~\cite{kipf2022conditional} poses the multi-object problem as optical-flow prediction using sequential frames as input. 
The internal slot-attention mechanism drives the network to learn regions that move in a simple and consistent manner. 
Different to our work, it does not assume a specific motion model but relies on directly regressing the flow. 
It is computationally more expensive and struggles when only one or few frames are available. 

\paragraph{Unsupervised video object segmentation.}
Unsupervised video object segmentation (VOS) is a popular problem in computer vision ~\cite{faktor2014videonlc,tsai2016video, papazoglou2013fast,tokmakov2019motion,jain2017fusionseg, yang2021self-supervised, yang-loquercio2019unsupervised, li2018instance,lu2019see}, that focuses on extracting the most salient object in the scene.
Many of the approaches treat the problem as a motion segmentation task as the background typically shows a dominant motion independent of the salient object. 
Motion Grouping~\cite{yang2021self-supervised} employs the Slot Attention architecture to reconstruct optical flow from itself, avoiding appearance information entirely.
Another related line of work~\cite{torr1998geometric, mahendran2018self-supervised, meunier2022em-driven, choudhury+karazija2022guess} employs approximate motion models. These approaches rely on a point estimate of the motion model parameters. In contrast, we adopt a more principled probabilistic approach, placing a prior on the motion parameters and integrating them out. To deal with flow outliers that do not conform to a rigid motion model, \citet{mahendran2018self-supervised} use a histogram matching-based loss and GWM~\citep{choudhury+karazija2022guess} over-segments the scene relying on spectral clustering to produce a binary segmentation during inference. Instead, we model the noise in our formulation directly. 
Finally, \citet{meunier2022em-driven} rely on flow as input, limiting the method to videos only. 

\section{Method}\label{s:method}

Let a frame $\mathcal{I} \in \mathbb{R}^{3\times H\times W}$ of a video and its optical flow $\mathbf{f}\in \mathbb{R}^{2\times H\times W}$ be defined on the $H \times W$ lattice.
The optical flow is a local summary of the motion from one frame to the next.
We use it to supervise a network $\Phi$ that, given the (single) image $\mathcal{I}$ as input, predicts soft assignments of each pixel to up to $K$ different image regions, outputting an $H\times W$ collection of probability vectors $\Phi(\mathcal{I})\in \hat \Delta_K^{H\times W} \subset [0,1]^{K\times H\times W}$, where $\hat \Delta_K$ is the $K-1$-dimensional simplex.
The quality of the regions is measured based on how likely they contain \textit{flow patterns} typical of the motion of independent objects.

In more detail, we represent the predicted image regions by a hard $K$-way pixel assignment (mask) $\mathbf{m} \in\Delta_K^{H\times W} \subset \{0,1\}^{K\times H\times W}$, where $\Delta_K$ is the space of $K$-dimensional one-hot vectors.
Each mask is a sample from the categorical distribution output by the network, \ie %
$
\mathbf{m} \sim p_\Phi(\mathbf{m}\mid\mathcal{I}) = \operatorname{Categorical}[\Phi(\mathcal{I})]
$.
Note that there is one categorical distribution for each pixel and that these are mutually independent.

We then assume that the flow depends only on the regions, in the sense that
$
p_\Phi(\mathbf{f},\mathbf{m}\mid\mathcal{I}) = p(\mathbf{f}\mid\mathbf{m})\, p_\Phi(\mathbf{m}\mid\mathcal{I}),
$
where $p(\mathbf{f}\mid\mathbf{m})$ is a model of the distribution of the flow field given the regions.
The likelihood of the modeled $\Phi$ is bounded by:
\begin{equation*}
\log p_\Phi(\mathbf{f} \mid\mathcal{I}) =
\log \mathop{\mathbb{E}}_{\mathbf{m}\sim p_\Phi(\mathbf{m}\mid\mathcal{I})}
\left[ p(\mathbf{f}\mid\mathbf{m}) \right]
\geq
\mathop{\mathbb{E}}_{\mathbf{m}\sim p_\Phi(\mathbf{m}\mid\mathcal{I})}
\left[ \log p(\mathbf{f}\mid\mathbf{m}) \right].
\end{equation*}
Furthermore, inspired by ELBO, we regularize the model's prediction $p_\Phi(\mathbf{m} \mid \mathcal{I})$ by taking its KL divergence from a uniform prior $p_0(\mathbf{m})$, obtaining the learning objective
\begin{equation}\label{e:nelbo}
\mathcal{L}_{\beta} = %
\displaystyle
\mathop{\mathbb{E}}_{\mathbf{m} \sim p_\Phi(\mathbf{m} \mid \mathcal{I})}
\left[-\log p(\mathbf{f}\mid\mathbf{m})\right] + \beta  D_{\text{KL}}\left(p_\Phi(\mathbf{m} \mid \mathcal{I})\,\Vert\,p_0(\mathbf{m})\right).
\end{equation}
Next, we introduce the closed-form motion model $p(\mathbf{f}\mid\mathbf{m})$ in~\cref{e:nelbo} and then explain how the Gumbel-Softmax trick can be used to train the network.

\paragraph{Approximate motion models for optical flow.}

We now turn to describing the models of motion used in our work, which play a role in assessing the likelihood of optical flow $p(\mathbf{f}\mid\mathbf{m})$.
Optical flow measures the coordinate change of pixels between neighboring frames, 
which arises due to the motion of the camera and objects.
We consider rigid-body motion of some object $k$.

Let
$
\mathbf{x}^{t}_k,\mathbf{y}^{t}_k \in \mathbb{R}^{n_k}
$
be the spatial locations of the pixels belonging to region/object $k$ at time $t$, where $n_k$ is the number of pixels in the region.
For convenience, we stack the coordinates in a single vector
$
\Omega^{t}_k = (\mathbf{x}^{t}_k,\mathbf{y}^{t}_k) \in \mathbb{R}^{2n_k}.
$
The pixels comprising this object undergo coordinate change from $\Omega^{t}_k$ to $\Omega^{t+1}_k$, giving rise to the optical flow for this object as
$
    \mathbf{f}_k = \Omega^{t+1}_k - \Omega^{t}_k.
$
We assume this underlying 3D rigid-body motion can be approximated using a linear 2D parametric model $\Pi_\theta$ with parameters $\theta$, so that:
\begin{equation}\label{e:motion_model}
    \Omega^{t+1}_k = \Pi_\theta(\Omega^{t}_k) + \epsilon,
    \qquad
    \mathbf{f}_k = \Pi_\theta(\Omega^{t}_k) - \Omega^{t}_k + \epsilon,
\end{equation}
where $\epsilon$ captures the residual error of the approximation. Several forms of models are available (see \citep{adiv1985determining,bergen1992hierarchical} for an overview).
Here, we consider two such models: the translation of an object within the camera plane, and an affine motion, given respectively by linear functions:
\begin{equation}
\Pi_\theta^{\mathrm{tr}}(\Omega_k^t)
= \Omega_k^t +
\underbrace{\begin{bmatrix}
    \mathbf{1}_{n_k} & 0 \\
    0 & \mathbf{1}_{n_k} \\
\end{bmatrix}}_{P_k^{\mathrm{tr}}}
\begin{pmatrix}\theta_1 \\ \theta_2 \end{pmatrix},
\quad
\Pi_{\theta}^{\mathrm{aff}}(\Omega_k^t)
=
\underbrace{\begin{bmatrix}
\mathbf{x}_{t} & \mathbf{y}_{t} & \mathbf{1}_{n_k}& 0 & 0 & 0 \\
0 & 0 & 0 & \mathbf{x}_{t} & \mathbf{y}_{t} & \mathbf{1}_{n_k}\end{bmatrix}}_{P_k^{\mathrm{aff}}}
\begin{pmatrix}\theta_1 \\ \vdots \\ \theta_6 \end{pmatrix},
\end{equation}
where we use $\mathbf{1}_{n_k}$ is a vector of $n_k$ ones and matrix $P_k$ contains the coefficients of the model.

The affine model supports object rotation, scaling and shearing in addition to translation.
It is often a sufficient approximation to real-world optical flow, provided the objects are rigid, convex, and mainly rotating in-plane.

We can then use the motion equations~\eqref{e:motion_model} to construct the distribution $p(\mathbf{f}\mid\mathbf{m})$ by assuming a prior on the motion parameters and by marginalizing over it.
Specifically, denote by $\mathbf{m}_k$ the $k$-th slice of the tensor $\mathbf{m}$ encoding the regions (\ie the mask of the $k$-th region).
We assume that regions are statistically independent and decompose the log-likelihood $p(\mathbf{f}\mid\mathbf{m})$ as:
\begin{equation}\label{e:basic}
    \log p(\mathbf{f}\mid\mathbf{m})
    = \sum_k \log p(\mathbf{f}_k\mid\mathbf{m}_k)
    = \sum_k \int \log p(\mathbf{f}_k, \theta_k\mid\mathbf{m}_k)d\theta_k.
\end{equation}
Assuming that each object has i.i.d.~parameters $\theta_k$ with a Gaussian prior
$
\mathcal{N}(\theta;\mu,\Sigma),
$
and assuming $\epsilon$ is a zero-mean noise with variance $\sigma^2$, \cref{e:motion_model} gives marginal optical flow likelihood for segment $k$:
\begin{equation}\label{e:flow_marginal}
p(\mathbf{f}_k\mid\mathbf{m}_k)
=
\mathcal{N}\left( \mathbf{f}_k;\,\Pi_\mu(\Omega_k) - \Omega_k,\, P_k\Sigma P_k^\top + \sigma^2I \right)
\end{equation}
where $I$ is the identity matrix. A practical issue with~\cref{e:flow_marginal} is that, if segment $k$ contains $n_k = \sum_i (\mathbf{m}_k)_i$ pixels, then the covariance matrix $P_k\Sigma P_k^\top + \sigma^2I$ has dimension $2n_k \times 2n_k$.
Inverting such a matrix in the evaluation of the Gaussian log-density is very slow except for very small regions.
Furthermore, it is not obvious how to relax~\cref{e:basic} to support gradient-based learning, \eg through the Gumbel-Softmax approximation.
We solve these problems in the next section.

\paragraph{Expressions for the likelihood.}

We now derive expressions for \cref{e:flow_marginal} which are efficient and that lead to a natural relaxation for use in the Gumbel-Softmax sampling.
Given the definitions
$
\mathbf{F}_k = \mathbf{f}_k - \Pi_\mu(\Omega_k) + \Omega_k$ and $\Lambda = \Sigma^{-1}
$,
we can rewrite~\cref{e:flow_marginal} as:
\begin{equation}\label{e:ops}
p(\mathbf{f}_k\mid\mathbf{m}_k)
=
(2\pi\sigma^2)^{-n_k}
\left(
    \frac
    {\det S_k}
    {\det \Lambda}
\right)^{-\frac{1}{2}}
e^{-\frac{d^2}{2\sigma^2}},
~~~
d^2=
\mathbf{F}_k^\top
\mathbf{F}_k - \frac{1}{\sigma^2} \mathbf{F}_k^\top
P_k S_k^{-1} P_k^\top\mathbf{F}_k \ ,
\end{equation}
where
$
S_k = \nicefrac{1}{\sigma^2}P_k^\top P_k + \Lambda.
$
The significant advantage of this form is that it involves the computation of the inverse and determinant of matrix $S_k$, whose size is only $2\times 2$ (for the translation model) or $6\times 6$ (for the affine one), instead of the much larger ${2n_k}\times {2n_k}$.

We can more explicitly introduce the dependency on the region assignments $\mathbf{m}$ by defining selector matrices $R_k \in \{0,1\}^{2n_k \times 2n}$ (with $n = \sum_k n_k = HW$) that extract the $\mathbf{x}$ and $\mathbf{y}$ coordinates of the pixels that belong to the corresponding region, \ie
$
\Omega_k = R_k \Omega.
$
We can then also write $\mathbf{F}_k = R_k \mathbf{F}$ and $P_k = R_k P$.
Furthermore, the product of the selectors
$
L_k = R_k^\top R_k \in \{0,1\}^{2n\times 2n}
$
can be written directly as a function of the assignment $\mathbf{m}$ as
$
L_k(\mathbf{m}) = \operatorname{diag}(\mathbf{m}_k, \mathbf{m}_k).
$
Plugging these back in~\cref{e:ops}, we obtain expressions involving $L_k$ only:
\begin{equation}
\label{e:diffme}
n_k
=
\frac{1}{2} |L_k|_1,
\quad
S_k
=
\frac{1}{\sigma^2}P^\top L_k P + \Lambda,
\quad
d^2
=
\mathbf{F}^\top L_k \mathbf{F}
- \frac{1}{\sigma^2}
(\mathbf{F}^\top L_k P)
S^{-1}_k
(P^\top L_k \mathbf{F}).
\end{equation}

\paragraph{Translation-only model.}

Further simplifications are possible for specific models.
For instance, for the translation-only model, assuming that
$
\Lambda = \operatorname{diag}(1/\tau^2, 1/\tau^2)
$
then
$
S_k = \operatorname{diag}(n_k + 1/\tau^2, n_k + 1/\tau^2)
$
and, after some calculations, we obtain the expression:
\begin{equation*}
- \log p(\mathbf{f} \mid \mathbf{m})
=
n \log 2\pi\sigma^2
+
\sum_k
\log \frac{n_k + \frac{\sigma^2}{\tau^2}}{\frac{\sigma^2}{\tau^2}}
+
\frac{1}{2\sigma^2}
\mathbf{F}^\top
\left(
I -
\sum_k
\frac{1}{n_k + \frac{\sigma^2}{\tau^2}}
\begin{bsmallmatrix}
    \mathbf{m}_k\mathbf{m}_k^\top &  \\
     & \mathbf{m}_k\mathbf{m}_k^\top \\
\end{bsmallmatrix}
\right)
\mathbf{F}.
\end{equation*}

\paragraph{Affine model.}

For the affine model, the expression for $-\log p (\mathbf{f}\mid\mathbf{m})$ does not simplify as much.
Still, by exploiting the structure of matrix $P_k^\mathrm{aff}$, we can reduce the calculations to the computation of inverse and determinant of small $3\times 3$ matrices, which can be implemented efficiently in closed form. 
Please see the Appendix for the derivation. 
Unless otherwise stated, the mean vector is set to $\mu = (1~0~0~0~1~0)^\top$ centering the prior on the no-motion point.

\paragraph{Gumbel-softmax.}

In order to train the network using gradient descent, we need a differentiable version of loss~\eqref{e:nelbo}.
To do so, we use the re-parametrizable Gumbel-softmax relaxation~\cite{maddison2016concrete,jang}.
The Gumbel-softmax relaxation replaces categorical samples $\mathbf{m} \in\Delta_K^{H\times W}$ from the distribution $p_\Phi(\mathbf{m}\mid\mathcal{I}) = \operatorname{Categorical}[\Phi(\mathcal{I})]$ with continuous samples $\hat{\mathbf{m}} \in \hat \Delta_K^{H\times W}$ from the distribution $\operatorname{GumbelSoftmax}[\Phi(\mathcal{I})]$.
We take $N=3$ samples from this distribution to evaluate the expected negative log-likelihood, further reducing variance.
Then we simply replace $\hat{\mathbf{m}}$ for $\mathbf{m}$ in \cref{e:nelbo}, leading to differentiable quantities.

\paragraph{Post-processing.}

\cref{e:flow_marginal} naturally encourages the model to form larger regions to explain parts of the scene that move in a consistent (under the assumed prior) manner.
However, we find that this can also lead to the model grouping together objects that only coincidentally move together (\eg all objects mostly falling due to gravity in one of the datasets).
Furthermore, optical flow is ambiguous around object edges and occlusions.
To address both the object grouping and occlusion boundary issue, we use a simple post-processing step.
We isolate connected components in the model output, selecting the $K$ largest masks, discarding any that are smaller than 0.1\% of the image area, and combining the left-over and discarded ones with the largest mask overall.

\paragraph{Warp loss.}
Occasionally, the optical flow used to supervise our model can be noisy as it is estimated by other methods. This noise is also unlikely to be isotropic as some surfaces are easier to estimate that others.
Rather than supporting heterogeneous noise and approximation error (\cref{e:motion_model}), we instead prioritize parts of the scene covered by higher-quality flow.
To this end, we introduce an additional loss term that simply enforces consistency between adjacent frames $\mathcal{I}_1, \mathcal{I}_2$.
In particular, it warps the predicted mask distributions $\Phi(\mathcal{I}_1), \Phi(\mathcal{I}_2)$ using the optical flow, weighted by the error of warping the frames themselves, as follows:
\begin{align}\label{e:warp_loss}
\mathcal{L}_{\mathrm{warp}}(\mathcal{I}_1, \mathcal{I}_2, f_1, b_2) &= w(\mathcal{I}_2, f_1(\mathcal{I}_1)) \cdot d(\Phi(\mathcal{I}_2), f_1(\Phi(\mathcal{I}_1))) \\
&+ w(\mathcal{I}_1, b_2(\mathcal{I}_2)) \cdot d(\Phi(\mathcal{I}_1), b_2(\Phi(\mathcal{I}_2))),\nonumber\\
w(\mathcal{I}_a, \mathcal{I}_b) &= 1 - \mathrm{norm}(|\mathcal{I}_a - \mathcal{I}_b|),\nonumber \\
d(p, q) &= \kldiv{p}{q}/2 + \kldiv{q}{p}/2,\nonumber
\end{align}
where $f_1(\cdot)$ indicates warping by forward optical flow $f_1$ (or backward $b_2$). The symmetrized KL divergence, $d(\cdot)$, measured agreement between predicted and warped mask distributions, weighted by the absolute error of the warped frames normalized in $[0, 1]$.
While the use of this term is not central to our method, we include it to show how tolerance to noisy optical flow can be improved.
We do not use the warp loss (\cref{e:warp_loss}) in our experiments, unless otherwise indicated by (WL). In that case, the final loss is simply sum of the two terms: $\mathcal{L}_{\beta} + \mathcal{L}_{\mathrm{warp}}$.

\section{Experiments}\label{s:experiments}

Our method lies in between image-based and video-based segmentation approaches, because it uses videos for supervision, but trains an image segmentation network that operates on still images only.
We thus evaluate our approach under a number of settings.
Firstly, we evaluate how well motion can be used to supervise an object instance segmenter that operates on still images.
Secondly, we compare such a segmenter to state-of-the-art object segmentation methods that use motion also at test time (and are thus advantaged compared to our model).
We conduct further analysis to validate our modelling assumptions and the model's reliance on the quality of the optical flow used for supervision.
Finally, we apply our method to a real-world setting.

\subsection{Experimental setup}\label{s:exp_setup}

\paragraph{Datasets.}\label{s:exp_setup:datasets}
We evaluate our method on video and still image datasets.
For video-based data, we use the Multi-Object Video (MOVi) datasets, released as part of Kubric~\cite{greff2021kubric}.
Specifically, we employ MOVi-\{A,C,D,E\} versions.
MOVi-A is similar to CLEVR~\cite{johnson2017clevr} in terms of visual complexity and contains videos of 3--10 falling objects on a simple, gray background.
MOVi-C is significantly more challenging, as it features scanned, textured, common objects on top of backgrounds textured using HDR images. In MOVi-D, the number of objects is increased up to 23. In MOVi-E, the camera is additionaly moving.
We use a resolution of $128 \times 128$ and the provided ground truth optical flow.

\newcommand{\imwidth}{0.095\linewidth}

\begin{figure}[t]
\centering
\minipage{0.97\linewidth}
\minipage{\imwidth}
  \includegraphics[width=\linewidth]{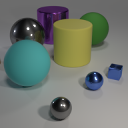}
\endminipage\hfill
\minipage{\imwidth}
  \includegraphics[width=\linewidth]{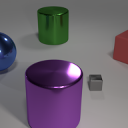}
\endminipage\hfill
\minipage{\imwidth}
  \includegraphics[width=\linewidth]{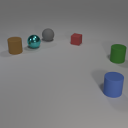}
\endminipage\hfill
\minipage{\imwidth}
  \includegraphics[width=\linewidth]{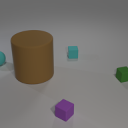}
\endminipage\hfill
\minipage{\imwidth}
    \includegraphics[width=\linewidth]{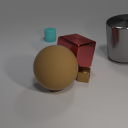}
\endminipage\hfill
\minipage{\imwidth}
    \includegraphics[width=\linewidth]{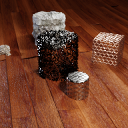}
\endminipage\hfill
\minipage{\imwidth}
    \includegraphics[width=\linewidth]{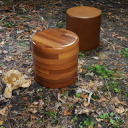}
\endminipage\hfill
\minipage{\imwidth}
    \includegraphics[width=\linewidth]{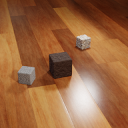}
\endminipage\hfill
\minipage{\imwidth}
    \includegraphics[width=\linewidth]{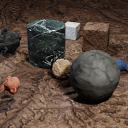}
\endminipage\hfill
\minipage{\imwidth}
    \includegraphics[width=\linewidth]{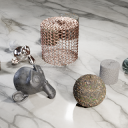}
\endminipage\hfill
\minipage{\imwidth}
  \includegraphics[width=\linewidth]{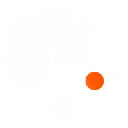}
\endminipage\hfill
\minipage{\imwidth}
    \includegraphics[width=\linewidth]{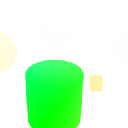}
\endminipage\hfill
\minipage{\imwidth}
    \includegraphics[width=\linewidth]{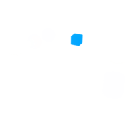}
\endminipage\hfill
\minipage{\imwidth}
    \includegraphics[width=\linewidth]{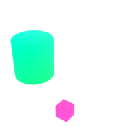}
\endminipage\hfill
\minipage{\imwidth}
    \includegraphics[width=\linewidth]{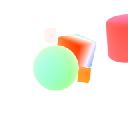}
\endminipage\hfill
\minipage{\imwidth}
    \includegraphics[width=\linewidth]{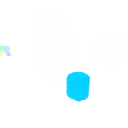}
\endminipage\hfill
\minipage{\imwidth}
    \includegraphics[width=\linewidth]{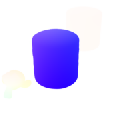}
\endminipage\hfill
\minipage{\imwidth}
    \includegraphics[width=\linewidth]{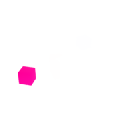}
\endminipage\hfill
\minipage{\imwidth}
    \includegraphics[width=\linewidth]{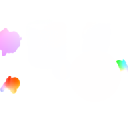}
\endminipage\hfill
\minipage{\imwidth}
    \includegraphics[width=\linewidth]{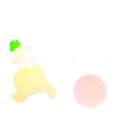}
\endminipage\hfill
\endminipage
\caption{Example images and optical flow of the \textsc{MovingCLEVR} and \textsc{MovingClevrTex} datasets, which extend \textsc{CLEVR} and \textsc{ClevrTex} respectively to short videos based on physics simulation. Note that only a subset of objects is in motion in each frame.}
\label{fig:moving_clevr_example}
\end{figure}

To evaluate our method on still images, we use \textsc{CLEVR}~\cite{johnson2017clevr} and \textsc{ClevrTex}~\cite{karazija2021clevrtex} benchmark suites.
Both consist of images depicting 3--10 objects.
\textsc{CLEVR} images are simpler with uniformly colored objects with metallic or rubbery materials.
\textsc{ClevrTex} features more diverse objects with complex textures applied.
We also use the \textsc{OOD} and \textsc{CAMO} test sets from \textsc{ClevrTex} benchmark.
\textsc{OOD} contains out-of-distribution shapes and textures.
\textsc{CAMO} has camouflaged objects where the same texture is sampled for objects and the background.

Since our method requires optical flow during training, we extend the implementation of~\cite{karazija2021clevrtex} to generate video datasets of \textsc{CLEVR} and \textsc{ClevrTex} scenes, where a \emph{subset} of objects contained in each scene are sliding, rolling and colliding based on a physics simulation (\cref{fig:moving_clevr_example}).
We generate 10k sequences for \textsc{MovingClevrTex} and 5k for \textsc{MovingCLEVR}, where we retain 1000 and 500 sequences, respectively, for validation.
Each sequence is 5 frames long.
Dataset details can be found in the Appendix.
The evaluation is performed on the original \textsc{CLEVR} and \textsc{ClevrTex} test sets.

We also evaluate our method on the real-world KITTI~\cite{geiger2012we} benchmark which depicts street scenes captured from a moving car.
We follow the set up of \cite{bao2022discovering}, using 147 videos for training and evaluate on the instance segmentation subset which contains 200 annotated validation frames. 

\paragraph{Metrics.} Following prior work~\cite{kipf2022conditional, karazija2021clevrtex}, we measure performance using two metrics.
FG-ARI is the Adjusted Rand Index measured on foreground pixels only (selected using the ground-truth segmentation).
Mean Intersection-over-Union (mIoU), is measured through Hungarian matching and averaged across the number of predicted or ground truth components, whichever is higher.
When evaluating on videos, we calculate these metrics per-frame. %

\paragraph{Network architecture.}
Our method can employ any image segmentation network architecture and train from scratch. 
Unless otherwise specified, we use Mask2Former~\citep{cheng2021mask2former}, using only its semantic segmentation. 
Following prior work~\citep{locatello2020object,kipf2022conditional}, we use a 6-layer CNN backbone on synthetic datasets and ResNet-18 for KITTI.
We also experiment with Swin-tiny transformer~\citep{liu2021swin} as the backbone.  
We use 11 transformer queries which become $K=11$ slots on \textsc{CLEVR}, \textsc{ClevrTex}, and MOVi-A/C. On MOVi-D/E, we set $K=24$ and $K=22$ on KITTI. 
The model takes approximately 48h to train on a single A30 24GB GPU.\footnote{Approx. total compute in this paper: 100 GPU days for our models, 154 GPU days for comparisons.}
All training details and hyper-parameters are included in the Appendix.

\subsection{Unsupervised multi-object segmentation in images}

\newcommand{\arif}[2]{
\tablenum[table-format=2.2]{#1}\small\color{gray}$\pm$\tablenum[table-format=2.2]{#2}
}
\newcommand{\arifbf}[2]{
\hspace{0.15em}\textbf{#1}\small\color{gray}$\pm$\tablenum[table-format=2.2]{#2}
}

\newcommand{\arifs}[2]{
\tablenum[table-format=2.2]{#1}\small\color{gray}$\pm$\tablenum[table-format=1.2]{#2}
}
\newcommand{\arifsbf}[2]{
\hspace{0.15em}\textbf{#1}\small\color{gray}$\pm$\tablenum[table-format=1.2]{#2}
}

\newcommand{\arifnoerr}[1]{
\tablenum[table-format=2.2]{#1}\hspace{3.1em}
}

\newcommand{\arifsnoerr}[1]{
\tablenum[table-format=2.2]{#1}\hspace{2.7em}
}

\newcommand{\mious}[2]{
\tablenum[table-format=2.2]{#1}\small\color{gray}$\pm$\tablenum[table-format=1.2]{#2}
}
\newcommand{\miousbf}[2]{
\hspace{0.05em}\textbf{#1}\small\color{gray}$\pm$\tablenum[table-format=1.2]{#2}
}

\newcommand{\miousnoerr}[1]{
\tablenum[table-format=2.2]{#1}\hspace{2.3em}
}

\newcommand{\mioud}[2]{
\tablenum[table-format=2.2]{#1}\small\color{gray}$\pm$\tablenum[table-format=2.2]{#2}
}
\newcommand{\mioudbf}[2]{
\hspace{0.05em}\textbf{#1}\small\color{gray}$\pm$\tablenum[table-format=2.2]{#2}
}

\begin{table}
\caption{Benchmark results on \textsc{CLEVR}, \textsc{ClevrTex}, \textsc{CAMO}, and \textsc{OOD} comparing FG-ARI and mIoU metrics (see also Appendix for an extended version). Results are a mean of 3 seeds $(\pm \sigma)$. Methods above the line are trained on single images, while methods below train on videos.\gray{$^\dagger$ -- indicates post-processing.}}
\label{tab:results_cle_clt_aris}
\resizebox{\textwidth}{!}{
\begin{tabular}{lrrrrrrrr}
\toprule
 & \multicolumn{2}{c}{\textsc{CLEVR}}
      & \multicolumn{2}{c}{\textsc{ClevrTex}} 
      & \multicolumn{2}{c}{\textsc{OOD}}
      & \multicolumn{2}{c}{\textsc{CAMO}}
\\
\cmidrule(l{2pt}r{2pt}){2-3}
\cmidrule(l{2pt}r{2pt}){4-5}
\cmidrule(l{2pt}r{2pt}){6-7}
\cmidrule(l{2pt}){8-9}

\textbf{Model}
& \multicolumn{1}{c}{\textbf{FG-ARI}\(\uparrow\)}
& \multicolumn{1}{c}{\textbf{mIoU}\(\uparrow\)}
& \multicolumn{1}{c}{\textbf{FG-ARI}\(\uparrow\)}
& \multicolumn{1}{c}{\textbf{mIoU}\(\uparrow\)}
& \multicolumn{1}{c}{\textbf{FG-ARI}\(\uparrow\)}
& \multicolumn{1}{c}{\textbf{mIoU}\(\uparrow\)}
& \multicolumn{1}{c}{\textbf{FG-ARI}\(\uparrow\)}
& \multicolumn{1}{c}{\textbf{mIoU}\(\uparrow\)}
\\
\midrule

SPAIR \citep{crawford2019spatially}
& \arif{77.13}{1.92} & \mioud{65.95}{4.02}
& \arif{0.00}{0.00} & \mious{0.00}{0.00}
& \arif{0.00}{0.00} & \mious{0.00}{0.00}
& \arif{0.00}{0.00} & \mious{0.00}{0.00} 
\\
SPAIR$^{\dagger}$
& \arif{77.05}{1.96} & \mioud{66.87}{9.65}
& \arif{0.00}{0.00} & \mious{0.00}{0.00}
& \arif{0.00}{0.00} & \mious{0.00}{0.00}
& \arif{0.00}{0.00} & \mious{0.00}{0.00}
\\

MN \citep{smirnov2021marionette}
& \arif{72.12}{0.64} & \mious{56.81}{0.40}
& \arif{38.31}{0.70} & \mious{10.46}{0.10}
& \arif{37.29}{1.04} & \mious{12.13}{0.19} 
& \arif{31.52}{0.87} & \mious{8.79}{0.15}
\\
MN$^{\dagger}$
& \arif{72.08}{0.62} & \mious{57.61}{0.40}
& \arif{38.34}{0.73} & \mious{10.34}{0.12}
& \arif{37.28}{1.07} & \mious{11.97}{0.21}
& \arif{31.54}{0.87} & \mious{8.77}{0.18}
\\

MONet \citep{burgess2019monet} 
& \arif{54.47}{11.41} & \mioud{30.66}{14.87}
& \arif{36.66}{0.87} & \mious{19.78}{1.02}
& \arif{32.97}{1.00} & \mious{19.30}{0.37}
& \arif{12.44}{0.73} & \mious{10.52}{0.38}
\\
MONet$^{\dagger}$
& \arif{61.36}{7.33} & \mioud{45.61}{4.80}
& \arif{35.64}{1.17} & \mious{23.59}{0.29}
& \arif{31.51}{1.46} & \mious{23.04}{0.52}
& \arif{9.94}{0.50}  & \mious{11.31}{0.30}
\\

SA \citep{locatello2020object}
& \arif{95.89}{2.37} & \mioud{36.61}{24.83}
& \arif{62.40}{2.23} & \mious{22.58}{2.07}
& \arif{58.45}{1.87} & \mious{20.98}{1.59}
& \arif{57.54}{1.01} & \mious{19.83}{1.41}
\\
SA$^{\dagger}$
& \arif{94.88}{1.67} & \mioud{37.68}{26.56}
& \arif{61.60}{2.29} & \mious{21.96}{1.79}
& \arif{57.41}{1.92} & \mious{20.60}{1.45}
& \arif{56.85}{1.12} & \mious{19.42}{1.42}
\\

IODINE \citep{greff2019iodine}
& \arif{93.81}{0.76} & \mioud{45.14}{17.85}
& \arif{59.52}{2.20} & \mious{29.17}{0.75}
& \arif{53.20}{2.55} & \mious{26.28}{0.85}
& \arif{36.31}{2.57} & \mious{17.52}{0.75}
\\
IODINE$^{\dagger}$
& \arif{93.68}{0.83} & \mioud{44.20}{18.67}
& \arif{60.63}{2.50} & \mious{29.40}{1.10}
& \arif{54.92}{2.24} & \mious{27.96}{0.81}
& \arif{38.29}{1.40} & \mious{18.87}{0.52}
\\

DTI-S \citep{monnier2021dtisprites}
& \arif{89.54}{1.44} & \mioud{48.74}{2.17}
& \arif{79.90}{1.37} & \mious{33.79}{1.30}
& \arif{73.67}{0.98} & \mious{32.55}{1.08}
& \arif{72.90}{1.89} & \mious{27.54}{1.55}
\\
DTI-S$^{\dagger}$
& \arif{89.86}{1.78} & \mioud{53.38}{3.51}
& \arif{79.86}{1.36} & \mious{32.20}{1.49}
& \arif{73.60}{0.97} & \mious{30.74}{1.22}
& \arif{72.89}{1.88} & \mious{26.30}{1.57}
\\

GNM \citep{jiang2020generative}
& \arif{65.05}{4.19} & \mioud{59.92}{3.72}
& \arif{53.37}{0.67} & \mious{42.25}{0.18}
& \arif{48.43}{0.86} & \mious{40.84}{0.30} 
& \arif{15.73}{0.89} & \mious{17.56}{0.74}
\\
GNM$^{\dagger}$ 
& \arif{65.67}{4.23} & \mioud{63.38}{3.76}
& \arif{53.38}{0.67} & \mious{44.30}{0.19}
& \arif{48.44}{0.86} & \mious{42.87}{0.28}
& \arif{15.72}{0.89} & \mious{18.53}{0.75}
\\

\midrule

SAVi~\citep{kipf2022conditional} 
& \multicolumn{1}{c}{---} & \multicolumn{1}{c}{---}
& \arifnoerr{49.54} & \miousnoerr{31.88} 
& \arifnoerr{42.68} & \miousnoerr{30.31} 
& \arifnoerr{42.67} & \miousnoerr{29.60} 
\\
Ours
& \arif{91.69}{0.30} & \mioud{66.70}{0.32}
& \arif{90.80}{0.22} & \mious{55.07}{0.44}
& \arif{76.01}{0.56} & \mious{46.84}{0.20}
& \arif{72.78}{1.31} & \mious{42.30}{1.09}
\\

Ours $^{\dagger}$
& \arifbf{95.94}{0.43} & \mioudbf{84.86}{4.06}
& \arifbf{92.61}{0.22} & \miousbf{77.67}{0.25}
& \arifbf{78.24}{0.43} & \miousbf{55.54}{0.44}
& \arifbf{77.43}{0.86} & \miousbf{56.43}{0.80}
\\

\bottomrule
\end{tabular}
}
\end{table}
\begin{figure}[t]
\centering
    \includegraphics[width=0.99\linewidth]{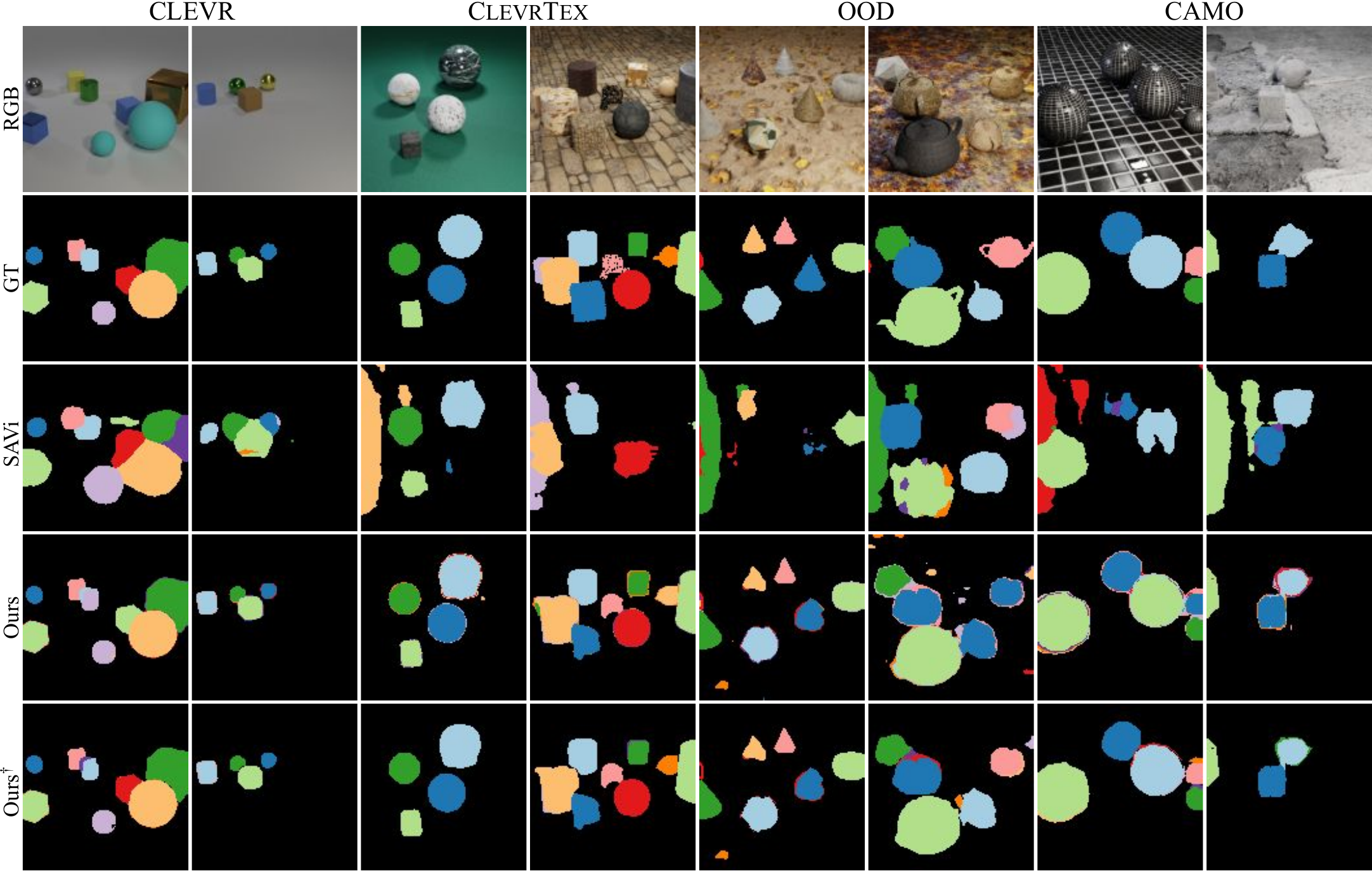}
    \caption{Unsupervised object segmentation on \textsc{CLEVR} and \textsc{ClevrTex} benchmarks. Our model is able to segment simple and visually complex scenes. Occasional mistakes around object boundaries and the assignment of different objects to the same component are addressed by post-processing. \gray{$^\dagger$ -- indicates post-processing.}}
    \label{fig:clevrtex_qual}
\end{figure}

In \cref{tab:results_cle_clt_aris}, we evaluate our method on the \textsc{CLEVR}~\citep{johnson2017clevr} and \textsc{ClevrTex}~\citep{karazija2021clevrtex} benchmarks and compare to prior work.
Our method outperforms image models based on appearance reconstruction on both metrics (mIoU and FG-ARI) and across all datasets.
The performance gap increases on the visually complex \textsc{ClevrTex}, \textsc{OOD}, and \textsc{CAMO} variants, demonstrating the strong inductive bias that motion provides during training, especially when the objects are camouflaged.
Note that, in this setting, our model is advantaged compared to the other models in \cref{tab:results_cle_clt_aris}, as it can observe (through the loss) the optical flow of the training scenes.
For this reason, we also train the optical flow-based, unconditional SAVi~\cite{kipf2022conditional} model.
Nevertheless, we find that despite having access to motion information during training, SAVi does not surpass appearance-only models, likely due to only having access to single frames at test time. 

Post-processing helps improve results further. 
As shown in \cref{fig:clevrtex_qual}, separating connected components in post-processing distinguishes objects that might be assigned to the same mask and suppresses boundary segments that tend to group difficult occlusion boundaries.
For a fairer comparison, we also test if this post-processing improves the results of other methods but only obtain mixed results.

\subsection{Unsupervised multi-object segmentation in video}
\begin{table}[t]
\centering \small
\caption{Segmentation results on MOVi datasets. Mean $\pm$ standard error (5 seeds). We calculate metric for each frame. All values in $\%$. (WL) marks use of warp loss. \gray{$^\dagger$ -- indicates post-processing.}}
\label{table:seg-results}
\resizebox{0.95\textwidth}{!}{
\begin{tabular}{lcccccccc} \\
\toprule
 & \multicolumn{2}{c}{MOVi-A} & \multicolumn{2}{c}{MOVi-C} 
 & \multicolumn{2}{c}{MOVi-D} & \multicolumn{2}{c}{MOVi-E} \\
 
\cmidrule(l{2pt}r{2pt}){2-3}
\cmidrule(l{2pt}r{2pt}){4-5}
\cmidrule(l{2pt}r{2pt}){6-7}
\cmidrule(l{2pt}r{2pt}){8-9}

\textbf{Model}& \textbf{FG-ARI}$\uparrow$ & \textbf{mIoU}$\uparrow$  & \textbf{FG-ARI}$\uparrow$ & \textbf{mIoU}$\uparrow$ & \textbf{FG-ARI}$\uparrow$ & \textbf{mIoU}$\uparrow$ & \textbf{FG-ARI}$\uparrow$ & \textbf{mIoU}$\uparrow$ \\
\midrule
GWM~\cite{choudhury+karazija2022guess} &  $70.30$ & $42.27$ & $49.98$ & $30.17$ & $39.78$ & $18.38$ & $42.50$ & $18.74$ \\

SCALOR~\citep{jiang2020scalor} & $59.57$ & $44.41$ & $40.43$ & $22.54$ & -- & -- & -- & -- \\ 
SAVi~\citep{kipf2022conditional} & $\mathbf{88.30}$ & $62.69$ & $43.26$ & $31.92$ & $43.45$ & $10.60$ & $17.39$ & $5.75$ \\
\midrule
Ours 
& \arifs{84.01}{0.72} & \mious{60.08}{1.47} 
& \arifs{61.18}{0.84} & \mious{34.72}{0.17} 
& \arifsbf{55.74}{1.02} & \mious{23.50}{0.35} 
& \arifs{62.62}{0.92} & \mious{25.78}{0.27} \\

Ours$^\dagger$ 
& \arifs{85.41}{1.00} & \miousbf{76.19}{2.05} 
& \arifsbf{61.24}{0.85} & \miousbf{37.26}{0.33} 
& \arifs{55.18}{0.94} & \miousbf{25.21}{0.29} 
& \arifsbf{63.11}{0.91} & \miousbf{28.59}{0.29} \\

\midrule
\midrule
Ours$^\dagger$ (Swin + WL)& $90.08$ & $84.76$ & $67.64$ & $52.17$ & $66.41$ & $30.40$ & $72.73$ & $35.30$ \\
\bottomrule
\end{tabular}
}
\label{table:main_quantitative_results}
\end{table}
  
\begin{figure}[t]
    \centering
    \includegraphics[page=1,width=0.99\linewidth,trim={0 4.2cm 0 0},clip]{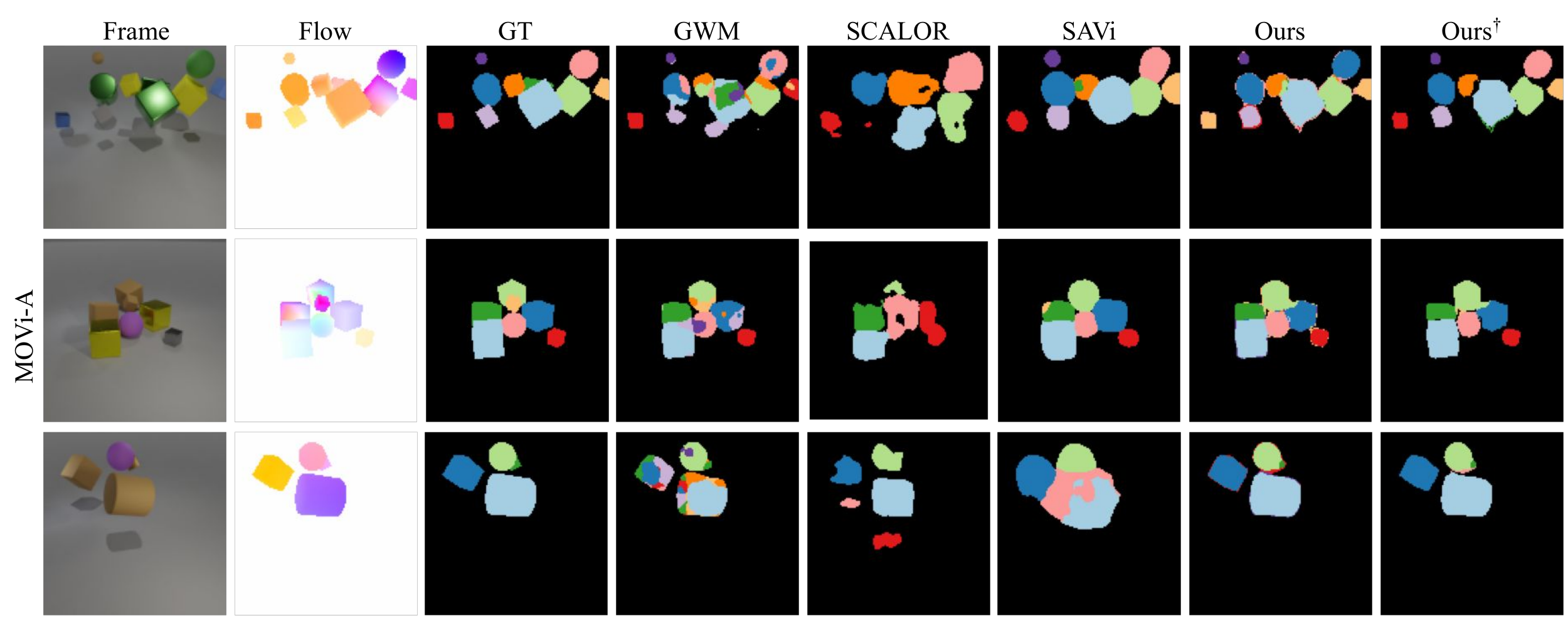} 
    \includegraphics[page=2,width=0.99\linewidth,trim={0 4.2cm 0 0.8cm},clip]{figures/images/neurips22_viz.pdf} 
    \caption{Qualitative comparisons on \movia and \movic.
     Our method performs consistently well compared to other methods. GWM suffers from oversegmentation where SCALOR has undersegmentation issue. %
     Among the related methods, SCALOR fails to discover all the objects and SAVi's object boundaries are coarser
     \gray{${}^\dagger$-- indicates post-processing.}}
    \label{fig:qualitative_video_movi}
\end{figure}

\begin{figure}[t]
    \centering
    \includegraphics[page=3,width=\linewidth,trim={0 7.5cm 0 0},clip]{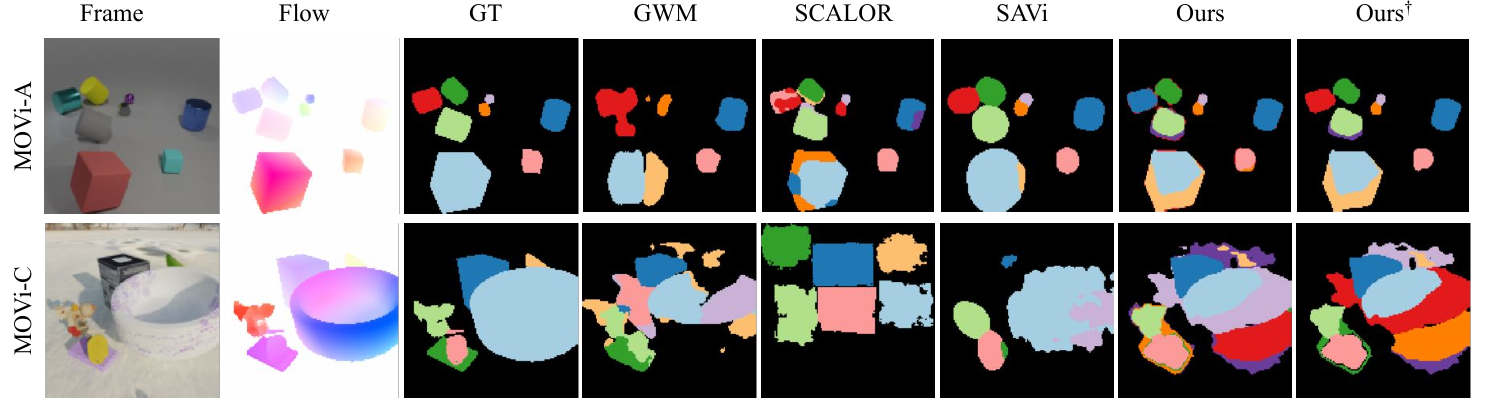} 
    \caption{Failure cases on \movia and \movic.
    Our method has difficulty with objects that typically exhibit complex motion. It results in an over-segmentation of the object. Due to inherently imprecise optical flow near the boundaries our method also has a tendency to segment boundary pixels into a separate mask, which we fix with our post-processing step. \gray{${}^\dagger$-- indicates post-processing.}}
    \label{fig:qualitative_video_failure}
\end{figure}

We now evaluate our approach on video segmentation, where motion is available at test time.
We report the performance of our method in \Cref{table:seg-results} compared to video-based models: SCALOR~\citep{jiang2020scalor}, \emph{unconditional} SAVi~\citep{kipf2022conditional}, and GWM~\citep{choudhury+karazija2022guess}; the latter two also use optical flow supervision.
It is important to note that SAVi and SCALOR make predictions jointly over all frames of a video, that allows them to actually see the objects in motion at test time.  
Despite this comparison being unfair to our approach, which operates on a single frame at a time, we achieve competitive results.  
On the visually simpler \movia, our method has more than 10\% lead over SAVi in mIoU, but performs slightly worse in terms of FG-ARI. 
On the more complex MoVi-C/D/E datasets our model shows strong performance, outperforming prior work on both metrics, with the performance gap again increasing with data complexity.
Finally, we experiment with a version of the model using a deeper backbone (Swin) and the warp loss (\cref{e:warp_loss}). Though not necessary to achieve state-of-art results, this drastically improves performance on all metrics and dataset versions.
In \cref{fig:qualitative_video_movi}, we also compare these models qualitatively, demonstrating that the different objects are overall better captured by our model with more refined boundaries, which explains the higher mIoU.

\subsection{Model ablations}

\paragraph{Optical flow.}

In \cref{tab:s:abl-flow}, we replace the ground-truth optical flow used so far in our experiments with the one estimated by SMURF~\cite{stone2021SMURFSM}.
The additional noise 
impacts our model's accuracy, %
as evidenced by the significant drop in mIoU (but comparable FG-ARI scores).

\paragraph{Motion model.}

In \cref{tab:s:abl-motion}, we compare our affine motion model to the simpler translation-only one.
We observe that the ability to describe complex motion patterns with an affine model improves performance over the translation model, which can only represent translation in the camera plane.

\begin{table}[t]
\centering
\caption{Model ablations.  (\ref{tab:s:abl-flow}) replacing flow supervision with an unsupervised flow method, (\ref{tab:s:abl-motion}) compares a translation-only with the affine flow model, and (\ref{tab:s:abl-loss}) adds our motion-based formulation to an appearance-only model. All models with post-processing applied.}
\begin{subtable}{0.46\linewidth}
\caption{Choice of Optical Flow Method}
\label{tab:s:abl-flow}
\vspace{-0.8em}
\resizebox{\textwidth}{!}{
\begin{tabular}{lcccc} \\
\toprule
 & \multicolumn{2}{c}{MOVi-A} & \multicolumn{2}{c}{MOVi-C} \\
\cmidrule(l{2pt}r{2pt}){2-3}\cmidrule(l{2pt}r{2pt}){4-5}
\textbf{Optical Flow}& \textbf{FG-ARI}$\uparrow$ & \textbf{mIoU}$\uparrow$  & \textbf{FG-ARI}$\uparrow$ & \textbf{mIoU}$\uparrow$ \\
\midrule
SMURF~\cite{stone2021SMURFSM} &  $80.17$ & $26.3$ & $61.21$ & $28.77$ \\
Ground Truth  &  $83.48$ & $72.61$ & $58.59$ & $35.67$ \\
\bottomrule
\end{tabular}
}
\end{subtable}
\hspace{0.01\linewidth}
\begin{subtable}{0.46\linewidth}
\caption{Choice of Motion Model}
\label{tab:s:abl-motion}
\vspace{-0.8em}
\resizebox{\textwidth}{!}{
\begin{tabular}{lcccc} \\
\toprule
 & \multicolumn{2}{c}{MOVi-A} & \multicolumn{2}{c}{MOVi-C} \\
\cmidrule(l{2pt}r{2pt}){2-3}\cmidrule(l{2pt}r{2pt}){4-5}
\textbf{Motion Mdl.}& \textbf{FG-ARI}$\uparrow$ & \textbf{mIoU}$\uparrow$  & \textbf{FG-ARI}$\uparrow$ & \textbf{mIoU}$\uparrow$ \\
\midrule
Translation &  $66.03$ & $59.94$ & $39.77$ & $32.23$ \\
Affine &  $83.48$ & $72.61$ & $58.59$ & $35.67$ \\
\bottomrule
\end{tabular}
}
\end{subtable}
\begin{subtable}{0.9\linewidth}
\caption{Adding motion awareness to appearance-only models}
\label{tab:s:abl-loss}
\resizebox{\textwidth}{!}{
\begin{tabular}{lrcccccc}
    \toprule
        &  & \multicolumn{2}{c}{\textsc{ClevrTex}} 
      & \multicolumn{2}{c}{\textsc{OOD}}
      & \multicolumn{2}{c}{\textsc{CAMO}}
\\
\cmidrule(l{2pt}r{2pt}){3-4}
\cmidrule(l{2pt}r{2pt}){5-6}
\cmidrule(l{2pt}){7-8}

\textbf{Model}
& \textbf{Train data}
& \multicolumn{1}{c}{\textbf{FG-ARI}\(\uparrow\)}
& \multicolumn{1}{c}{\textbf{mIoU}\(\uparrow\)}
& \multicolumn{1}{c}{\textbf{FG-ARI}\(\uparrow\)}
& \multicolumn{1}{c}{\textbf{mIoU}\(\uparrow\)}
& \multicolumn{1}{c}{\textbf{FG-ARI}\(\uparrow\)}
& \multicolumn{1}{c}{\textbf{mIoU}\(\uparrow\)}
\\
\midrule

GNM \citep{jiang2020generative} & \textsc{ClevrTex}
& $53.38$ & $44.30$
& $48.44$ & $42.87$
& $15.72$ & $18.53$
\\

GNM \citep{jiang2020generative} & \textsc{MovingClevrTex}
& $18.01$ & $31.47$
& $15.57$ & $15.57$
& $0.21$ & $14.68$
\\

GNM+Our Loss & \textsc{MovingClevrTex}
& $63.84$ & $55.26$ 
& $59.01$ & $48.65$ 
& $51.00$ & $47.63$ 
\\

\midrule
SA \citep{locatello2020object} & \textsc{ClevrTex}
& $62.40$ & $22.58$
& $58.45$ & $20.98$
& $57.54$ & $19.83$
\\

SA \citep{locatello2020object} & \textsc{MovingClevrTex}
& $61.84$ & $21.44$
& $58.24$ & $20.67$
& $57.30$ & $18.82$
\\

SA+Our Loss & \textsc{MovingClevrTex}
& $76.60$ & $38.12$
& $67.01$ & $33.95$
& $70.59$ & $33.05$
\\

\bottomrule
\end{tabular}
}
\end{subtable}

\label{table:ablations_1}
\end{table}
\paragraph{Motion awareness.}

Finally, we investigate the effectiveness of our objective in combination with existing appearance-based methods.
The goal of this experiment is to understand the advantage of using motion information during training (if available) and to decouple the effect of our formulation from the choice of architecture. 
To this end, we employ our objective on top of two models based on appearance reconstruction, GNM~\cite{jiang2020generative} and SA~\cite{locatello2020object}, with no other modifications.
We train GNM and SA with and without our loss on videos (\textsc{MovingClevrTex}) and evaluate on the corresponding single-image test sets of the \textsc{ClevrTex} suite.
In \cref{tab:s:abl-loss}, we compare these models respectively to the original methods trained on static images (\textsc{ClevrTex}). 
We find that when trained on video data (\textsc{MovingClevrTex}), without our loss, both GNM and SA struggle. 
We attribute this to the reduced number of scenes in \textsc{MovingClevrTex} compared to \textsc{ClevrTex}.
However, we note that using our loss significantly improves the performance of the appearance methods, %
suggesting the effectiveness of exploiting motion information through our formulation.

\subsection{Segmentation on real-world data}

\begin{wraptable}[16]{r}{4.5cm}
\centering
\vspace{-1.3em}
\caption{Real-world segmentation results on KITTI. Baseline results from \cite{bao2022discovering}. \citet{bao2022discovering} and our method use RAFT for optical flow. Models above the line use ResNet-18 backbone. (WL) marks use of warp loss.}
\label{tab:kitti}
\resizebox{0.25\textwidth}{!}{
\begin{tabular}{lc}
\toprule
& KITTI \\
\cmidrule(l{2pt}r{2pt}){2-2}
\textbf{Model} & \textbf{FG-ARI}$\uparrow$ \\
\midrule
SA \cite{locatello2020object} & $13.8$\\
MONet \cite{burgess2019monet} & $14.9$\\
SCALOR \cite{liu2020learning} & $21.1$\\
S-IODINE \cite{greff2019iodine} & $14.4$\\
MCG \cite{arbelaez2014multiscale} & $40.9$ \\
\citet{bao2022discovering} & $47.1$ \\
\textbf{Ours} & $50.8$ \\
\textbf{Ours} (WL) & $\mathbf{51.9}$ \\
\midrule
\textbf{Ours} (Swin + WL) & $58.3$ \\
\bottomrule
\end{tabular}%
}%
\end{wraptable}%
We now turn to assessing our model's performance in a real-world setting.
We follow the setting of \citet{bao2022discovering} and evaluate on KITTI~\cite{geiger2012we}, using RAFT~\cite{teed2020raft} to estimate optical flow and ResNet-18 as the backbone of our model, trained from scratch.
We lower the input resolution for our model from $368 \times 1,248$~\citep{bao2022discovering} to $288 \times 960$, which enables us to fit on a single GPU. We evaluate at $96 \times 320$ resolution.

As we show in \cref{tab:kitti}, our method outperforms prior work on the challenging real-world setting.
In the same table, we also consider our method with an additional warp loss term, which further boosts performance.  We also experiment with a transformer-based backbone (Swin) which is also pre-trained using self-supervision.
Although, not necessary to show state-of-art result, this significantly improves real-world performance. 

\subsection{Limitations}
\label{s:limitations}
Motion is sometimes insufficient to distinguish different objects, for instance because they do not move or because they move similarly.
In principle, this should not matter if sufficient motion diversity is observed in the training data as a whole; in practice, our model occasionally merges different object at test time, which we address partly in post-processing.
Further improvements could be obtained by choosing a more informative prior $p_0(\mathbf{m})$ in~\cref{e:nelbo}, to capture other desirable properties of objects, such as compactness and connectedness.

\Cref{fig:qualitative_video_failure} shows some failure cases where the affine motion model struggles to capture strong perspective effects caused by non-smooth depth changes in the object geometry.
This could be addressed by modeling the object geometry (depth) and the ensuing complex flow patterns. Alternatively, this could be dealt with using a hierarchical segmentation model that can account for geometric discontinuities and self-occlusions.
Hierarchical segmentation would also help with pronounced non-rigid motion (\eg humans dancing or animals running) as motion could be explained at the level of object parts.

\section{Conclusions}\label{s:conclusions}

We have presented a method that bridges the gap between image-based and video-based scene decomposition, in that it requires only a single image as input, yet exploits motion cues available in videos during training. 
In comparison to prior work on image-based multi-object segmentation, our approach shows that motion provides useful objectness cues, especially as the visual complexity of a scene increases.  
Different from video-based approaches, however, our model operates on still images and does not rely on motion to detect or refine objects, which makes it more generally applicable.  
Finally, we deviate from the common objective of image or flow \emph{reconstruction} and, instead, model the problem by only predicting regions likely to contain affine flow patterns.
This does not require a specialized architecture, thus any segmentation network is suitable for this task.  
Our approach achieves state-of-the-art performance on multiple image and video benchmarks, in simulated and real-world settings, validating the paradigm of training image models using motion.

\paragraph{Broader impact.}\label{s:impact}

Our work introduces a principled method for unsupervised multi-object segmentation.
The work is mainly evaluated on 3D simulated datasets that do not contain people or personal information. 
Additionally, we evaluate on KITTI, a real-world self-driving dataset, which  occasionally contains images of pedestrians.
Consent cannot be obtained in this case, but we follow the KITTI terms of usage.
We build on top of open source projects, respecting licenses and release all code, trained models and datasets for research purposes.
Currently, the application of this approach is mainly limited to simulated imagery. Although the results on KITTI show promise, the immediate broader impact of our work in real-world scenarios, beyond the research community, is limited. 
\paragraph{Acknowledgements}
L.~K. is funded by EPSRC Centre for Doctoral Training in Autonomous Intelligent Machines and Systems EP/S024050/1. 
S. C. is supported by a scholarship sponsored by Facebook. 
I.~L., C.~R. and A.~V. are supported by European Research Council (ERC) grant 2020-CoG-101001212 UNION.
I.~L. and C.~R. are also funded by EPSRC grant VisualAI EP/T028572/1.

\clearpage
{\small\bibliographystyle{plainnat}\bibliography{refs}}

\begin{thebibliography}{61}
\providecommand{\natexlab}[1]{#1}
\providecommand{\url}[1]{\texttt{#1}}
\expandafter\ifx\csname urlstyle\endcsname\relax
  \providecommand{\doi}[1]{doi: #1}\else
  \providecommand{\doi}{doi: \begingroup \urlstyle{rm}\Url}\fi

\bibitem[Adiv(1985)]{adiv1985determining}
Gilad Adiv.
\newblock Determining three-dimensional motion and structure from optical flow
  generated by several moving objects.
\newblock \emph{IEEE transactions on pattern analysis and machine
  intelligence}, \penalty0 (4):\penalty0 384--401, 1985.

\bibitem[Arbelaez et~al.(2014)Arbelaez, Pont-Tuset, Barron, Marques, and
  Malik]{arbelaez2014multiscale}
Pablo Arbelaez, Jordi Pont-Tuset, Jonathan~T. Barron, Ferran Marques, and
  Jitendra Malik.
\newblock Multiscale combinatorial grouping.
\newblock In \emph{Proceedings of the IEEE Conference on Computer Vision and
  Pattern Recognition (CVPR)}, June 2014.

\bibitem[Bao et~al.(2022)Bao, Tokmakov, Jabri, Wang, Gaidon, and
  Hebert]{bao2022discovering}
Zhipeng Bao, Pavel Tokmakov, Allan Jabri, Yu-Xiong Wang, Adrien Gaidon, and
  Martial Hebert.
\newblock Discovering objects that can move.
\newblock In \emph{Proceedings of the IEEE/CVF Conference on Computer Vision
  and Pattern Recognition (CVPR)}, pages 11789--11798, June 2022.

\bibitem[Bergen et~al.(1992)Bergen, Anandan, Hanna, and
  Hingorani]{bergen1992hierarchical}
James~R Bergen, Patrick Anandan, Keith~J Hanna, and Rajesh Hingorani.
\newblock Hierarchical model-based motion estimation.
\newblock In \emph{European conference on computer vision}, pages 237--252.
  Springer, 1992.

\bibitem[Burgess et~al.(2019)Burgess, Matthey, Watters, Kabra, Higgins,
  Botvinick, and Lerchner]{burgess2019monet}
Christopher~P Burgess, Loic Matthey, Nicholas Watters, Rishabh Kabra, Irina
  Higgins, Matt Botvinick, and Alexander Lerchner.
\newblock Monet: Unsupervised scene decomposition and representation.
\newblock \emph{arXiv preprint arXiv:1901.11390}, 2019.

\bibitem[Cheng et~al.(2022)Cheng, Misra, Schwing, Kirillov, and
  Girdhar]{cheng2021mask2former}
Bowen Cheng, Ishan Misra, Alexander~G Schwing, Alexander Kirillov, and Rohit
  Girdhar.
\newblock Masked-attention mask transformer for universal image segmentation.
\newblock In \emph{Proceedings of the IEEE/CVF Conference on Computer Vision
  and Pattern Recognition}, pages 1290--1299, 2022.

\bibitem[Choudhury et~al.(2021)Choudhury, Laina, Rupprecht, and
  Vedaldi]{choudhury2021unsupervised}
Subhabrata Choudhury, Iro Laina, Christian Rupprecht, and Andrea Vedaldi.
\newblock Unsupervised part discovery from contrastive reconstruction.
\newblock In \emph{Advances in Neural Information Processing Systems},
  volume~35, 2021.

\bibitem[Choudhury et~al.(2022)Choudhury, Karazija, Laina, Vedaldi, and
  Rupprecht]{choudhury+karazija2022guess}
Subhabrata Choudhury, Laurynas Karazija, Iro Laina, Andrea Vedaldi, and
  Christian Rupprecht.
\newblock Guess {W}hat {M}oves: {U}nsupervised video and image segmentation by
  anticipating motion.
\newblock In \emph{British Machine Vision Conference (BMVC)}, 2022.

\bibitem[Crawford and Pineau(2019)]{crawford2019spatially}
Eric Crawford and Joelle Pineau.
\newblock Spatially invariant unsupervised object detection with convolutional
  neural networks.
\newblock In \emph{Proceedings of the AAAI Conference on Artificial
  Intelligence}, volume~33, pages 3412--3420, 2019.

\bibitem[Dutt~Jain et~al.(2017)Dutt~Jain, Xiong, and
  Grauman]{jain2017fusionseg}
Suyog Dutt~Jain, Bo~Xiong, and Kristen Grauman.
\newblock Fusionseg: Learning to combine motion and appearance for fully
  automatic segmentation of generic objects in videos.
\newblock In \emph{Proceedings of the IEEE conference on computer vision and
  pattern recognition}, pages 3664--3673, 2017.

\bibitem[Emami et~al.(2021)Emami, He, Ranka, and
  Rangarajan]{emami2021efficient}
Patrick Emami, Pan He, Sanjay Ranka, and Anand Rangarajan.
\newblock Efficient iterative amortized inference for learning symmetric and
  disentangled multi-object representations.
\newblock In \emph{Proceedings of the 38th International Conference on Machine
  Learning}, pages 2970--2981. PMLR, 2021.

\bibitem[Engelcke et~al.(2020)Engelcke, Kosiorek, Jones, and
  Posner]{engelcke2019genesis}
Martin Engelcke, Adam~R Kosiorek, Oiwi~Parker Jones, and Ingmar Posner.
\newblock Genesis: Generative scene inference and sampling with object-centric
  latent representations.
\newblock In \emph{International Conference on Learning Representations}, 2020.

\bibitem[Engelcke et~al.(2021)Engelcke, Jones, and Posner]{engelcke2021genesis}
Martin Engelcke, Oiwi~Parker Jones, and Ingmar Posner.
\newblock Genesis-v2: Inferring unordered object representations without
  iterative refinement.
\newblock In \emph{Advances in Neural Information Processing Systems},
  volume~34, 2021.

\bibitem[Eslami et~al.(2016)Eslami, Heess, Weber, Tassa, Szepesvari,
  Kavukcuoglu, and Hinton]{eslami2016attend}
S.~M.~Ali Eslami, Nicolas Heess, Theophane Weber, Yuval Tassa, David
  Szepesvari, Koray Kavukcuoglu, and Geoffrey~E. Hinton.
\newblock Attend, infer, repeat: Fast scene understanding with generative
  models.
\newblock In \emph{Proceedings of the 30th International Conference on Neural
  Information Processing Systems}, page 3233–3241, 2016.

\bibitem[Faktor and Irani(2014)]{faktor2014videonlc}
Alon Faktor and Michal Irani.
\newblock Video segmentation by non-local consensus voting.
\newblock In \emph{Proceedings of the British Machine Vision Conference}. BMVA
  Press, 2014.

\bibitem[Geiger et~al.(2012)Geiger, Lenz, and Urtasun]{geiger2012we}
Andreas Geiger, Philip Lenz, and Raquel Urtasun.
\newblock Are we ready for autonomous driving? {T}he {KITTI} vision benchmark
  suite.
\newblock In \emph{Proceedings of the IEEE/CVF Conference on Computer Vision
  and Pattern Recognition (CVPR)}, pages 3354--3361. IEEE, 2012.

\bibitem[Greff et~al.(2015)Greff, Srivastava, and
  Schmidhuber]{Greff2015BindingVR}
Klaus Greff, Rupesh~Kumar Srivastava, and J{\"u}rgen Schmidhuber.
\newblock Binding via reconstruction clustering.
\newblock \emph{ArXiv}, abs/1511.06418, 2015.

\bibitem[Greff et~al.(2016)Greff, Rasmus, Berglund, Hao, Valpola, and
  Schmidhuber]{greff2016tagger}
Klaus Greff, Antti Rasmus, Mathias Berglund, Tele Hao, Harri Valpola, and
  J{\"u}rgen Schmidhuber.
\newblock Tagger: Deep unsupervised perceptual grouping.
\newblock In \emph{Advances in Neural Information Processing Systems}, pages
  4484--4492, 2016.

\bibitem[Greff et~al.(2019)Greff, Kaufman, Kabra, Watters, Burgess, Zoran,
  Matthey, Botvinick, and Lerchner]{greff2019iodine}
Klaus Greff, Rapha{\"e}l~Lopez Kaufman, Rishabh Kabra, Nick Watters,
  Christopher Burgess, Daniel Zoran, Loic Matthey, Matthew Botvinick, and
  Alexander Lerchner.
\newblock Multi-object representation learning with iterative variational
  inference.
\newblock In \emph{International Conference on Machine Learning}, pages
  2424--2433. PMLR, 2019.

\bibitem[Greff et~al.(2022)Greff, Belletti, Beyer, Doersch, Du, Duckworth,
  Fleet, Gnanapragasam, Golemo, Herrmann, Kipf, Kundu, Lagun, Laradji, Liu,
  Meyer, Miao, Nowrouzezahrai, Oztireli, Pot, Radwan, Rebain, Sabour, Sajjadi,
  Sela, Sitzmann, Stone, Sun, Vora, Wang, Wu, Yi, Zhong, and
  Tagliasacchi]{greff2021kubric}
Klaus Greff, Francois Belletti, Lucas Beyer, Carl Doersch, Yilun Du, Daniel
  Duckworth, David~J Fleet, Dan Gnanapragasam, Florian Golemo, Charles
  Herrmann, Thomas Kipf, Abhijit Kundu, Dmitry Lagun, Issam Laradji,
  Hsueh-Ti~(Derek) Liu, Henning Meyer, Yishu Miao, Derek Nowrouzezahrai, Cengiz
  Oztireli, Etienne Pot, Noha Radwan, Daniel Rebain, Sara Sabour, Mehdi S.~M.
  Sajjadi, Matan Sela, Vincent Sitzmann, Austin Stone, Deqing Sun, Suhani Vora,
  Ziyu Wang, Tianhao Wu, Kwang~Moo Yi, Fangcheng Zhong, and Andrea
  Tagliasacchi.
\newblock Kubric: a scalable dataset generator.
\newblock In \emph{Proceedings of the IEEE Conference on Computer Vision and
  Pattern Recognition (CVPR)}, 2022.

\bibitem[He et~al.(2019)He, Li, Liu, He, and Barber]{He2019TrackingBA}
Zhen He, Jian Li, Daxue Liu, Hangen He, and David Barber.
\newblock Tracking by animation: Unsupervised learning of multi-object
  attentive trackers.
\newblock \emph{2019 IEEE/CVF Conference on Computer Vision and Pattern
  Recognition (CVPR)}, pages 1318--1327, 2019.

\bibitem[Jang et~al.(2017)Jang, Gu, and Poole]{jang}
Eric Jang, Shixiang Gu, and Ben Poole.
\newblock Categorical reparameterization with gumbel-softmax.
\newblock In \emph{5th International Conference on Learning Representations,
  {ICLR}}, 2017.

\bibitem[Jiang and Ahn(2020)]{jiang2020generative}
Jindong Jiang and Sungjin Ahn.
\newblock Generative neurosymbolic machines.
\newblock In \emph{Advances in Neural Information Processing Systems},
  volume~33, pages 12572--12582, 2020.

\bibitem[Jiang et~al.(2020)Jiang, Janghorbani, Melo, and Ahn]{jiang2020scalor}
Jindong Jiang, Sepehr Janghorbani, Gerard~De Melo, and Sungjin Ahn.
\newblock Scalor: Generative world models with scalable object representations.
\newblock In \emph{International Conference on Learning Representations}, 2020.

\bibitem[Johnson et~al.(2017)Johnson, Hariharan, Van Der~Maaten, Fei-Fei,
  Lawrence~Zitnick, and Girshick]{johnson2017clevr}
Justin Johnson, Bharath Hariharan, Laurens Van Der~Maaten, Li~Fei-Fei,
  C~Lawrence~Zitnick, and Ross Girshick.
\newblock {CLEVR}: A diagnostic dataset for compositional language and
  elementary visual reasoning.
\newblock In \emph{Proceedings of the IEEE Conference on Computer Vision and
  Pattern Recognition}, pages 2901--2910, 2017.

\bibitem[Kabra et~al.(2021)Kabra, Zoran, Erdogan, Matthey, Creswell, Botvinick,
  Lerchner, and Burgess]{Kabra2021SIMONeVT}
Rishabh Kabra, Daniel Zoran, Goker Erdogan, Lo{\"i}c Matthey, Antonia Creswell,
  Matthew~M. Botvinick, Alexander Lerchner, and Christopher~P. Burgess.
\newblock {SIMON}e: {V}iew-invariant, temporally-abstracted object
  representations via unsupervised video decomposition.
\newblock volume~34, pages 20146--20159, 2021.

\bibitem[Karazija et~al.(2021)Karazija, Laina, and
  Rupprecht]{karazija2021clevrtex}
Laurynas Karazija, Iro Laina, and Christian Rupprecht.
\newblock Clevrtex: A texture-rich benchmark for unsupervised multi-object
  segmentation.
\newblock In \emph{Thirty-fifth Conference on Neural Information Processing
  Systems Datasets and Benchmarks Track}, 2021.

\bibitem[Kipf et~al.(2022)Kipf, Elsayed, Mahendran, Stone, Sabour, Heigold,
  Jonschkowski, Dosovitskiy, and Greff]{kipf2022conditional}
Thomas Kipf, Gamaleldin~Fathy Elsayed, Aravindh Mahendran, Austin Stone, Sara
  Sabour, Georg Heigold, Rico Jonschkowski, Alexey Dosovitskiy, and Klaus
  Greff.
\newblock Conditional object-centric learning from video.
\newblock In \emph{International Conference on Learning Representations}, 2022.

\bibitem[Kosiorek et~al.(2018)Kosiorek, Kim, Teh, and
  Posner]{kosiorek2018sequential}
Adam Kosiorek, Hyunjik Kim, Yee~Whye Teh, and Ingmar Posner.
\newblock Sequential attend, infer, repeat: Generative modelling of moving
  objects.
\newblock \emph{Advances in Neural Information Processing Systems}, 31, 2018.

\bibitem[Kossen et~al.(2020)Kossen, Stelzner, Hussing, Voelcker, and
  Kersting]{kossen2020structured}
Jannik Kossen, Karl Stelzner, Marcel Hussing, Claas Voelcker, and Kristian
  Kersting.
\newblock Structured object-aware physics prediction for video modeling and
  planning.
\newblock In \emph{Proceedings of the International Conference on Learning
  Representations}, 2020.

\bibitem[Li et~al.(2020)Li, Eastwood, and Fisher]{Li2020LearningOR}
Nanbo Li, Cian Eastwood, and Robert Fisher.
\newblock Learning object-centric representations of multi-object scenes from
  multiple views.
\newblock \emph{Advances in Neural Information Processing Systems},
  33:\penalty0 5656--5666, 2020.

\bibitem[Li et~al.(2018)Li, Seybold, Vorobyov, Fathi, Huang, and
  Kuo]{li2018instance}
Siyang Li, Bryan Seybold, Alexey Vorobyov, Alireza Fathi, Qin Huang, and
  C-C~Jay Kuo.
\newblock Instance embedding transfer to unsupervised video object
  segmentation.
\newblock In \emph{Proceedings of the IEEE conference on computer vision and
  pattern recognition}, pages 6526--6535, 2018.

\bibitem[Lin et~al.(2020{\natexlab{a}})Lin, Wu, Peri, Fu, Jiang, and
  Ahn]{Lin2020ImprovingGI}
Zhixuan Lin, Yi-Fu Wu, Skand Peri, Bofeng Fu, Jindong Jiang, and Sungjin Ahn.
\newblock Improving generative imagination in object-centric world models.
\newblock In \emph{International Conference on Machine Learning}, pages
  6140--6149. PMLR, 2020{\natexlab{a}}.

\bibitem[Lin et~al.(2020{\natexlab{b}})Lin, Wu, Peri, Sun, Singh, Deng, Jiang,
  and Ahn]{lin2020space}
Zhixuan Lin, Yi-Fu Wu, Skand~Vishwanath Peri, Weihao Sun, Gautam Singh, Fei
  Deng, Jindong Jiang, and Sungjin Ahn.
\newblock {SPACE}: Unsupervised object-oriented scene representation via
  spatial attention and decomposition.
\newblock In \emph{International Conference on Learning Representations},
  2020{\natexlab{b}}.

\bibitem[Liu et~al.(2020)Liu, Zhang, He, Liu, Wang, Tai, Luo, Wang, Li, and
  Huang]{liu2020learning}
Liang Liu, Jiangning Zhang, Ruifei He, Yong Liu, Yabiao Wang, Ying Tai, Donghao
  Luo, Chengjie Wang, Jilin Li, and Feiyue Huang.
\newblock Learning by analogy: Reliable supervision from transformations for
  unsupervised optical flow estimation.
\newblock In \emph{Proceedings of the IEEE/CVF Conference on Computer Vision
  and Pattern Recognition (CVPR)}, June 2020.

\bibitem[Liu et~al.(2021)Liu, Lin, Cao, Hu, Wei, Zhang, Lin, and
  Guo]{liu2021swin}
Ze~Liu, Yutong Lin, Yue Cao, Han Hu, Yixuan Wei, Zheng Zhang, Stephen Lin, and
  Baining Guo.
\newblock Swin transformer: Hierarchical vision transformer using shifted
  windows.
\newblock In \emph{Proceedings of the IEEE/CVF International Conference on
  Computer Vision}, pages 10012--10022, 2021.

\bibitem[Locatello et~al.(2020)Locatello, Weissenborn, Unterthiner, Mahendran,
  Heigold, Uszkoreit, Dosovitskiy, and Kipf]{locatello2020object}
Francesco Locatello, Dirk Weissenborn, Thomas Unterthiner, Aravindh Mahendran,
  Georg Heigold, Jakob Uszkoreit, Alexey Dosovitskiy, and Thomas Kipf.
\newblock Object-centric learning with slot attention.
\newblock In \emph{Advances in Neural Information Processing Systems},
  volume~33, pages 11525--11538, 2020.

\bibitem[Loshchilov and Hutter(2017)]{loshchilov2017decoupled}
Ilya Loshchilov and Frank Hutter.
\newblock Decoupled weight decay regularization.
\newblock \emph{arXiv preprint arXiv:1711.05101}, 2017.

\bibitem[Lu et~al.(2019)Lu, Wang, Ma, Shen, Shao, and Porikli]{lu2019see}
Xiankai Lu, Wenguan Wang, Chao Ma, Jianbing Shen, Ling Shao, and Fatih Porikli.
\newblock See more, know more: Unsupervised video object segmentation with
  co-attention siamese networks.
\newblock In \emph{Proceedings of the IEEE/CVF Conference on Computer Vision
  and Pattern Recognition}, pages 3623--3632, 2019.

\bibitem[Maddison et~al.(2017)Maddison, Mnih, and Teh]{maddison2016concrete}
Chris~J Maddison, Andriy Mnih, and Yee~Whye Teh.
\newblock The concrete distribution: A continuous relaxation of discrete random
  variables.
\newblock In \emph{5th International Conference on Learning Representations,
  {ICLR}}, 2017.

\bibitem[Mahendran et~al.(2018)Mahendran, Thewlis, and
  Vedaldi]{mahendran2018self-supervised}
Aravindh Mahendran, James Thewlis, and Andrea Vedaldi.
\newblock Self-supervised segmentation by grouping optical-flow.
\newblock In \emph{Proceedings of the European Conference on Computer Vision
  (ECCV) Workshops}, September 2018.

\bibitem[Meunier et~al.(2022)Meunier, Badoual, and
  Bouthemy]{meunier2022em-driven}
Etienne Meunier, Ana{\"\i}s Badoual, and Patrick Bouthemy.
\newblock {EM}-driven unsupervised learning for efficient motion segmentation.
\newblock \emph{arXiv preprint arXiv:2201.02074}, 2022.

\bibitem[Monnier et~al.(2021)Monnier, Vincent, Ponce, and
  Aubry]{monnier2021dtisprites}
Tom Monnier, Elliot Vincent, Jean Ponce, and Mathieu Aubry.
\newblock Unsupervised layered image decomposition into object prototypes.
\newblock In \emph{Proceedings of the IEEE/CVF International Conference on
  Computer Vision (ICCV)}, pages 8640--8650, 2021.

\bibitem[Papazoglou and Ferrari(2013)]{papazoglou2013fast}
Anestis Papazoglou and Vittorio Ferrari.
\newblock Fast object segmentation in unconstrained video.
\newblock In \emph{Proceedings of the IEEE International Conference on Computer
  Vision (ICCV)}, December 2013.

\bibitem[Ronneberger et~al.(2015)Ronneberger, Fischer, and
  Brox]{ronneberger2015u}
Olaf Ronneberger, Philipp Fischer, and Thomas Brox.
\newblock U-net: Convolutional networks for biomedical image segmentation.
\newblock In \emph{International Conference on Medical image computing and
  computer-assisted intervention}, pages 234--241. Springer, 2015.

\bibitem[Singh et~al.(2021)Singh, Peri, Kim, Kim, and
  Ahn]{Singh2021StructuredWB}
Gautam Singh, Skand Peri, Junghyun Kim, Hyunseok Kim, and Sungjin Ahn.
\newblock Structured world belief for reinforcement learning in {POMDP}.
\newblock In \emph{International Conference on Machine Learning}, pages
  9744--9755. PMLR, 2021.

\bibitem[Smirnov et~al.(2021)Smirnov, Gharbi, Fisher, Guizilini, Efros, and
  Solomon]{smirnov2021marionette}
Dmitriy Smirnov, Michael Gharbi, Matthew Fisher, Vitor Guizilini, Alexei~A
  Efros, and Justin Solomon.
\newblock Marionette: Self-supervised sprite learning.
\newblock In \emph{Advances in Neural Information Processing Systems},
  volume~34, 2021.

\bibitem[Stelzner et~al.(2019)Stelzner, Peharz, and Kersting]{stelzner19faster}
Karl Stelzner, Robert Peharz, and Kristian Kersting.
\newblock Faster attend-infer-repeat with tractable probabilistic models.
\newblock In \emph{Proceedings of the 36th International Conference on Machine
  Learning}, volume~97, pages 5966--5975. PMLR, 2019.

\bibitem[Stone et~al.(2021)Stone, Maurer, Ayvaci, Angelova, and
  Jonschkowski]{stone2021SMURFSM}
Austin Stone, Daniel Maurer, Alper Ayvaci, Anelia Angelova, and Rico
  Jonschkowski.
\newblock Smurf: Self-teaching multi-frame unsupervised raft with full-image
  warping.
\newblock \emph{2021 IEEE/CVF Conference on Computer Vision and Pattern
  Recognition (CVPR)}, pages 3886--3895, 2021.

\bibitem[Teed and Deng(2020)]{teed2020raft}
Zachary Teed and Jia Deng.
\newblock Raft: Recurrent all-pairs field transforms for optical flow.
\newblock In \emph{European Conference on Computer Vision (ECCV)}, 2020.

\bibitem[Tokmakov et~al.(2019)Tokmakov, Schmid, and
  Alahari]{tokmakov2019motion}
Pavel Tokmakov, Cordelia Schmid, and Karteek Alahari.
\newblock Learning to segment moving objects.
\newblock \emph{International Journal of Computer Vision}, 127\penalty0
  (3):\penalty0 282–301, mar 2019.

\bibitem[Torr(1998)]{torr1998geometric}
Philip H.~S. Torr.
\newblock Geometric motion segmentation and model selection.
\newblock \emph{Philosophical Transactions of the Royal Society of London.
  Series A: Mathematical, Physical and Engineering Sciences}, 356:\penalty0
  1321 -- 1340, 1998.

\bibitem[Tsai et~al.(2016)Tsai, Yang, and Black]{tsai2016video}
Yi-Hsuan Tsai, Ming-Hsuan Yang, and Michael~J. Black.
\newblock Video segmentation via object flow.
\newblock In \emph{2016 IEEE Conference on Computer Vision and Pattern
  Recognition (CVPR)}, pages 3899--3908, 2016.

\bibitem[Van~Steenkiste et~al.(2018)Van~Steenkiste, Chang, Greff, and
  Schmidhuber]{van2018relational}
Sjoerd Van~Steenkiste, Michael Chang, Klaus Greff, and J{\"u}rgen Schmidhuber.
\newblock Relational neural expectation maximization: Unsupervised discovery of
  objects and their interactions.
\newblock \emph{arXiv preprint arXiv:1802.10353}, 2018.

\bibitem[Veerapaneni et~al.(2020)Veerapaneni, Co-Reyes, Chang, Janner, Finn,
  Wu, Tenenbaum, and Levine]{veerapaneni2020entity}
Rishi Veerapaneni, John~D Co-Reyes, Michael Chang, Michael Janner, Chelsea
  Finn, Jiajun Wu, Joshua Tenenbaum, and Sergey Levine.
\newblock Entity abstraction in visual model-based reinforcement learning.
\newblock In \emph{Conference on Robot Learning}, pages 1439--1456. PMLR, 2020.

\bibitem[Weis et~al.(2021)Weis, Chitta, Sharma, Brendel, Bethge, Geiger, and
  Ecker]{Weis2021}
Marissa~A. Weis, Kashyap Chitta, Yash Sharma, Wieland Brendel, Matthias Bethge,
  Andreas Geiger, and Alexander~S. Ecker.
\newblock Benchmarking unsupervised object representations for video sequences.
\newblock \emph{Journal of Machine Learning Research}, 22\penalty0
  (183):\penalty0 1--61, 2021.

\bibitem[Wu et~al.(2021)Wu, Jones, Engelcke, and Posner]{Wu2021APEXUO}
Yizhe Wu, Oiwi~Parker Jones, Martin Engelcke, and Ingmar Posner.
\newblock Apex: Unsupervised, object-centric scene segmentation and tracking
  for robot manipulation.
\newblock \emph{2021 IEEE/RSJ International Conference on Intelligent Robots
  and Systems (IROS)}, pages 3375--3382, 2021.

\bibitem[Xu et~al.(2019)Xu, Li, Zhu, and Zhang]{kun2019deep}
Kun Xu, Chongxuan Li, Jun Zhu, and Bo~Zhang.
\newblock Multi-object generation with amortized structural regularization.
\newblock In \emph{Proceedings of the 33rd International Conference on Neural
  Information Processing Systems}. Curran Associates Inc., 2019.

\bibitem[Yang et~al.(2021)Yang, Lamdouar, Lu, Zisserman, and
  Xie]{yang2021self-supervised}
Charig Yang, Hala Lamdouar, Erika Lu, Andrew Zisserman, and Weidi Xie.
\newblock Self-supervised video object segmentation by motion grouping.
\newblock In \emph{Proceedings of the IEEE/CVF International Conference on
  Computer Vision}, pages 7177--7188, 2021.

\bibitem[Yang et~al.(2019)Yang, Loquercio, Scaramuzza, and
  Soatto]{yang-loquercio2019unsupervised}
Yanchao Yang, Antonio Loquercio, Davide Scaramuzza, and Stefano Soatto.
\newblock Unsupervised moving object detection via contextual information
  separation.
\newblock In \emph{Proceedings of the IEEE/CVF Conference on Computer Vision
  and Pattern Recognition (CVPR)}, June 2019.

\bibitem[Zablotskaia et~al.(2021)Zablotskaia, Dominici, Sigal, and
  Lehrmann]{zablotskaia21aPROVIDE}
Polina Zablotskaia, Edoardo~A. Dominici, Leonid Sigal, and Andreas~M. Lehrmann.
\newblock Provide: a probabilistic framework for unsupervised video
  decomposition.
\newblock In \emph{Proceedings of the Thirty-Seventh Conference on Uncertainty
  in Artificial Intelligence}, volume 161, pages 2019--2028. PMLR, 2021.

\end{thebibliography}

\clearpage

\section*{Checklist}
\begin{enumerate}

\item For all authors...
\begin{enumerate}
  \item Do the main claims made in the abstract and introduction accurately reflect the paper's contributions and scope?
    \answerYes{}
  \item Did you describe the limitations of your work?
    \answerYes{See \cref{s:limitations}.}
  \item Did you discuss any potential negative societal impacts of your work?
    \answerYes{See \cref{s:impact}}
  \item Have you read the ethics review guidelines and ensured that your paper conforms to them?
    \answerYes{}
\end{enumerate}

\item If you are including theoretical results...
\begin{enumerate}
  \item Did you state the full set of assumptions of all theoretical results?
    \answerYes{}
        \item Did you include complete proofs of all theoretical results?
    \answerYes{Longer form derivations are included in the supplementary material.}
\end{enumerate}

\item If you ran experiments...
\begin{enumerate}
  \item Did you include the code, data, and instructions needed to reproduce the main experimental results (either in the supplemental material or as a URL)?
    \answerYes{We include instruction and specification of our methods. See project page \url{https://www.robots.ox.ac.uk/~vgg/research/ppmp} for code and checkpoints.}
  \item Did you specify all the training details (e.g., data splits, hyperparameters, how they were chosen)?
    \answerYes{See supplementary material}
        \item Did you report error bars (e.g., with respect to the random seed after running experiments multiple times)?
    \answerYes{See \eg \cref{tab:results_cle_clt_aris} and \cref{table:main_quantitative_results}}
        \item Did you include the total amount of compute and the type of resources used (e.g., type of GPUs, internal cluster, or cloud provider)?
    \answerYes{See \cref{s:exp_setup}}
\end{enumerate}

\item If you are using existing assets (e.g., code, data, models) or curating/releasing new assets...
\begin{enumerate}
  \item If your work uses existing assets, did you cite the creators?
    \answerYes{See \cref{s:exp_setup:datasets}}
  \item Did you mention the license of the assets?
    \answerYes
  \item Did you include any new assets either in the supplemental material or as a URL?
    \answerYes{We provide the code for dataset creation and examples of the dataset in the supplementary material (due to size). See project page \url{https://www.robots.ox.ac.uk/~vgg/research/ppmp} for data.}
  \item Did you discuss whether and how consent was obtained from people whose data you're using/curating?
    \answerYes{}{We include one experiment on the KITTI dataset following their terms of usage. This data contains accidentally-captured pedestrians, from which consent cannot be obtained. These are ignored in the evaluation.}
  \item Did you discuss whether the data you are using/curating contains personally identifiable information or offensive content?
    \answerYes{See \cref{s:impact}.}
\end{enumerate}

\item If you used crowdsourcing or conducted research with human subjects...
\begin{enumerate}
  \item Did you include the full text of instructions given to participants and screenshots, if applicable?
    \answerNA{}
  \item Did you describe any potential participant risks, with links to Institutional Review Board (IRB) approvals, if applicable?
    \answerNA{}
  \item Did you include the estimated hourly wage paid to participants and the total amount spent on participant compensation?
    \answerNA{}
\end{enumerate}

\end{enumerate}

\clearpage  %
\appendix
\section*{Supplementary Material}
\label{s:supmat}
\setcounter{equation}{8} %

In this supplementary material, we provide additional details for our loss function (\cref{s:sup_loss_deriv}), including detailed derivation steps, implementation details, and further discussion of the advantages. \cref{s:sup_hparams} specifies hyperparameters used and how they were selected. We conclude with additional ablation experiments (\cref{s:sup_ablations}) and results (\cref{s:sup_qual_res}).
Project page and code: \url{https://www.robots.ox.ac.uk/~vgg/research/ppmp}.

\section{Loss derivation}\label{s:sup_loss_deriv}
The key part of our loss function is the likelihood of the optical flow $p(\mathbf{f}\mid\mathbf{m})$, which serves to evaluate how probable is a region of the optical flow carved out by the predicted masks for K regions. 
We assume that optical flow within a region depends only on the region itself and that other regions have no influence. Intuitively, this is a reasonable assumption, as in large, the movement/flow of an object does not depend on the background or other objects. Thus, enforcing this assumption encourages regions to correspond to objects.

The assessment of probability of optical flow within the region is based on the assumption that objects should be moving rigidly. We use an approximate parametric motion model (Eq. (2)). The parameters of the motion model $\theta$ abstract away unknown aspects such as scene geometry and camera intrinsics but enable to translate between assumed 3D rigid motion and 2D optical flow.

We assume that motion parameters $\theta_k \sim \mathcal{N}(\theta; \mu, \Sigma)$ come from a multivariate Gaussian prior. This choice enables expressing marginal-likelihood in closed-form.

We model the error of the approximate motion model as zero-mean isotropic Gaussian noise $\epsilon \sim \mathcal{N}(\epsilon; 0, \sigma^2 I)$. 

Following the motion model (Eq. (2)), the optical flow $\mathbf{f}_k$ is an \emph{affine combination} of Gaussian random variables. Using this observation, its distribution is 
\begin{equation}\label{e:sup:marginal}
    p(\mathbf{f}_k\mid\mathbf{m}_k) = \det(2\pi(P_k\Sigma P_k^\top + \sigma^2I))^{-\nicefrac{1}{2}} \cdot \exp(-\frac{1}{2}\mathbf{F}_k^\top(P_k\Sigma P_k^\top + \sigma^2I)^{-1}\mathbf{F}_k),
\end{equation}
where $\mathbf{F}_k = \mathbf{f}_k - \Pi_\mu(\Omega_k) + \Omega_k$ is the centered flow within the region $k$. This equation can be slightly simplified by considering its two troublesome parts, the determinant and the quadratic form inside the exponent. For the determinant, we note the following:
\begin{align*}
\det(2\pi P_k \Sigma P_k^\top + 2\pi\sigma^2I) &=
(2\pi\sigma^2)^{2n_k}\det(\nicefrac{1}{\sigma^2}P_k \Sigma P_k^\top + I) \\
&= (2\pi\sigma^2)^{2n_k}\det(\Sigma)\det(\nicefrac{1}{\sigma^2}P_k^\top P_k + \Sigma^{-1}) \tag{*}\\
&= \frac{(2\pi\sigma^2)^{2n_k}}{\det(\Lambda)}\det(\underbrace{\nicefrac{1}{\sigma^2}P_k^\top P_k + \Lambda}_{S_k}),
\end{align*}
where in the line marked with (*) we apply matrix determinant lemma, and in the last line we substitute covariance $\Sigma^{-1} = \Lambda$ for the precision matrix. 
Similarly, the quadratic form in the exponent can be expanded
\begin{align*}
\mathbf{F}_k^\top(P_k \Sigma P_k^\top + \sigma^2I)^{-1}\mathbf{F}_k &=
\nicefrac{1}{\sigma^2}\mathbf{F}_k^\top(\nicefrac{1}{\sigma^2}P_k \Sigma P_k^\top + I)^{-1}\mathbf{F}_k \\
&=\nicefrac{1}{\sigma^2}\mathbf{F}_k^\top\left(I - \nicefrac{1}{\sigma^2}P_k(\nicefrac{1}{\sigma^2}P_k^\top P_k + \Lambda)^{-1}P_k^\top\right)\mathbf{F}_k \tag{$\dagger$}\\
&=\nicefrac{1}{\sigma^2}\mathbf{F}_k^\top\mathbf{F}_k - (\nicefrac{1}{\sigma^2})^2\mathbf{F}_k^\top P_k(\underbrace{\nicefrac{1}{\sigma^2}P_k^\top P_k + \Lambda}_{S_k})^{-1}P_k^\top\mathbf{F}_k,
\end{align*}
where in the line marked with ($\dagger$) the Woodbury identity is applied. 
The optical flow for the whole image is modeled as a joint of independent flow regions $\mathbf{f}_k$, giving the log-likelihood as
\begin{align}\label{e:sup:log_exp_marginal}
& \log p(\mathbf{f}\mid\mathbf{m}) = \sum_k \log p(\mathbf{f}_k\mid\mathbf{m}_k) \nonumber\\ 
&= -\frac{1}{2}\left(2\log(2\pi\sigma^2) \sum_k n_k + \sum_k\log\frac{\det S_k}{\det \Lambda} + \nicefrac{1}{\sigma^2}\sum_k\mathbf{F}_k^\top\mathbf{F}_k - (\nicefrac{1}{\sigma^2})^2\sum_k\mathbf{F}_k^\top P_k S_k^{-1} P_k^\top\mathbf{F}_k\right) \nonumber\\
&= -\frac{1}{2}\left(2\log(2\pi\sigma^2)n + \sum_k\log\frac{\det S_k}{\det \Lambda} + \nicefrac{1}{\sigma^2}\mathbf{F}^\top\mathbf{F} - (\nicefrac{1}{\sigma^2})^2\sum_k\mathbf{F}^\top L_k P S_k^{-1} P^\top L_k \mathbf{F}\right),
\end{align}
where in the last line we introduced $n = \sum_k n_k = HW$, the number of pixels in the image. We also use product of selector matrices $L_k = R_k^\top R_k = \operatorname{diag}(\mathbf{m}_k,\mathbf{m}_k) = L_k ^ \top$ such that $\mathbf{F}_k^\top P_k = \mathbf{F}^\top L_k P$. This explicitly includes masks in the expression. Finally, we use the fact that regions partition the full image $\sum_k\mathbf{F}_k^\top\mathbf{F}_k = \sum_k \sum_i (\mathbf{F}_k)_i^2 =  \mathbf{F}^\top\mathbf{F}$. We now manipulate \cref{e:sup:log_exp_marginal} using specific details of the motion model to arrive at expressions that are convenient to implement in code.

\subsection{Implementation details}
\paragraph{Translation-only likelihood} 
We assume that translation along x and y directions is independent, such that $\theta$ prior is a zero-mean Gaussian with isotropic covariance $\tau^2I$. 
$P_k^{\mathrm{tr}} = \operatorname{diag}(\mathbf{1}_{n_k}, \mathbf{1}_{n_k})$ and by extension $P^{\mathrm{tr}} = \operatorname{diag}(\mathbf{1}_{n}, \mathbf{1}_{n})$. The matrix $S_k$ simplifies to
\begin{equation*}
    S_k = \nicefrac{1}{\sigma^2}P_k^\top P_k + \Lambda 
= \nicefrac{1}{\sigma^2}\begin{psmallmatrix}
    \mathbf{1}_{n_k}^\top \mathbf{1}_{n_k} & 0 \\
    0 & \mathbf{1}_{n_k}^\top \mathbf{1}_{n_k}
\end{psmallmatrix} + \nicefrac{1}{\tau^2}I 
= \nicefrac{1}{\sigma^2}(n_k +\nicefrac{\sigma^2}{\tau^2})I\,,  
\end{equation*}%
such that
\begin{equation*}
\det S_k = (\nicefrac{1}{\sigma^2}(n_k +\nicefrac{\sigma^2}{\tau^2}))^2,
\quad
\log\frac{\det S_k}{\det{\Lambda}} = 2\log\frac{n_k +\nicefrac{\sigma^2}{\tau^2}}{\nicefrac{\sigma^2}{\tau^2}},
\quad
S_k^{-1} = (\nicefrac{1}{\sigma^2}(n_k +\nicefrac{\sigma^2}{\tau^2}))^{-1}I.
\end{equation*}
Writing $\mathbf{F}^\top = (\mathbf{u}^\top, \mathbf{v}^\top)$ to denote x and y components of the flow, respectively, the term reduces
\begin{align*}
(\nicefrac{1}{\sigma^2})^2\sum_k\mathbf{F}^\top L_k P S_k^{-1} P^\top L_k \mathbf{F} &=
\nicefrac{1}{\sigma^2}\sum_k\mathbf{F}^\top L_k^\top P P^\top L_k \mathbf{F} \frac{1}{n_k +\nicefrac{\sigma^2}{\tau^2}} \\
&=
\nicefrac{1}{\sigma^2}\sum_k\frac{1}{n_k +\nicefrac{\sigma^2}{\tau^2}}
\begin{bsmallmatrix}
    \mathbf{u}^\top & \mathbf{v}^\top \\
\end{bsmallmatrix}
\begin{bsmallmatrix}
    \mathbf{m}_k\mathbf{m}_k^\top &  \\
     & \mathbf{m}_k\mathbf{m}_k^\top \\
\end{bsmallmatrix}
\begin{bsmallmatrix}
    \mathbf{u} \\
    \mathbf{v} \\
\end{bsmallmatrix}\\
&=\nicefrac{1}{\sigma^2}\sum_k\frac{1}{n_k +\nicefrac{\sigma^2}{\tau^2}}
\left(
(\mathbf{u}^\top\mathbf{m}_k)^2 + (\mathbf{v}^\top\mathbf{m}_k)^2
\right)\\
&=\nicefrac{1}{\sigma^2}\sum_k\frac{n_k^2}{n_k +\nicefrac{\sigma^2}{\tau^2}}
(\bar{u}_k^2 + \bar{v}_k^2)\, ,
\end{align*}
where in the last line we introduced mean flow $\bar{u}_k = n_k^{-1} \mathbf{u}^\top\mathbf{m}_k$ and $\bar{v}_k = n_k^{-1} \mathbf{v}^\top\mathbf{m}_k$. This gives the negative log-likelihood as
\begin{align}\label{e:sup:trans_marginal}
& \log p(\mathbf{f}\mid\mathbf{m}) = \sum_k \log p(\mathbf{f}_k\mid\mathbf{m}_k) \nonumber\\
&= -\frac{1}{2}\left(2\log(2\pi\sigma^2)n + \sum_k\log\frac{\det S_k}{\det \Lambda} + \nicefrac{1}{\sigma^2}\mathbf{F}^\top\mathbf{F} - (\nicefrac{1}{\sigma^2})^2\sum_k\mathbf{F}^\top L_k P S_k^{-1} P^\top L_k \mathbf{F}\right) \nonumber\\
&= -n\log(2\pi\sigma^2) - \sum_k\log\frac{n_k +\nicefrac{\sigma^2}{\tau^2}}{\nicefrac{\sigma^2}{\tau^2}} - \frac{1}{2\sigma^2}\left(\mathbf{F}^\top\mathbf{F} - \sum_k\frac{n_k^2(\bar{u}_k^2 + \bar{v}_k^2)}{n_k +\nicefrac{\sigma^2}{\tau^2}}\right) \nonumber\\
&=-n\log(2\pi\sigma^2) - \sum_k\log\frac{n_k +\nicefrac{\sigma^2}{\tau^2}}{\nicefrac{\sigma^2}{\tau^2}} - \frac{1}{2\sigma^2}\sum_{i=1}^{n}\left(u_i^2 + v_i^2 - \sum_k\frac{n_k(\bar{u}_k^2 + \bar{v}_k^2)}{n_k +\nicefrac{\sigma^2}{\tau^2}}(\mathbf{m}_k)_i\right).
\end{align}
In our implementation, we extend this equation further. Writing $w_k=1- \sqrt\frac{\nicefrac{\sigma^2}{\tau^2}}{n_k + \nicefrac{\sigma^2}{\tau^2}}$, we note that the following sum is equivalent

\begin{align*}
    \sum_{i=1}^n (u_i - \sum_k \bar{u}_k w_k (\mathbf{m}_k)_i)^2 &=
    \sum_{i=1}^n (u_i^2 - \sum_k 2 u_i \bar{u}_k w_k (\mathbf{m}_k)_i + \sum_k(\mathbf{m}_k)_i w_k^2\bar{u}_k^2)\\
&=
 \sum_{i=1}^n u_i^2 - \sum_k2\bar{u}_k w_k \sum_{i=1}^n u_i(\mathbf{m}_k)_i + \sum_kw_k^2\bar{u}_k^2\sum_{i=1}^n(\mathbf{m}_k)_i\\
&=
 \sum_{i=1}^n u_i^2 - \sum_k2\bar{u}_k^2 w_k n_k + \sum_k w_k^2\bar{u}_k^2 n_k\\
&=
 \sum_{i=1}^n u_i^2 - \sum_k\bar{u}_k^2 n_k(2w_k - w_k^2) \\
&=
 \sum_{i=1}^n u_i^2 - \sum_k\bar{u}_k^2\frac{n_k}{n_k + \nicefrac{\sigma^2}{\tau^2}}\sum_{i=1}^n(\mathbf{m}_k)_i \\
&=\sum_{i=1}^n \left(u_i^2 - \sum_k\frac{n_k\bar{u}_k^2(\mathbf{m}_k)_i}{n_k + \nicefrac{\sigma^2}{\tau^2}}\right),
\end{align*}

where in the first line we make use of the fact that masks are one-hot $(\mathbf{m})_i \in \{0,1\}^k$, thus only a single term in the sums over $k$ is non-zero, \ie $(\sum_kw_k\bar{u}_k(\mathbf{m}_k)_i)^2 = \sum_k w_k^2\bar{u}_k^2(\mathbf{m}_k)_i)$. Using the above insight, the log-likelihood is 
\begin{multline}\label{e:sup:trans_loglike}
    \log p(\mathbf{f}\mid\mathbf{m}) = -n\log(2\pi\sigma^2) - \sum_k\log\frac{n_k +\nicefrac{\sigma^2}{\tau^2}}{\nicefrac{\sigma^2}{\tau^2}} 
    \\- \frac{1}{2\sigma^2}\sum_{i=1}^{n}\left((u_i - \sum_k \bar{u}_k w_k (\mathbf{m}_k)_i)^2 + (v_i - \sum_k \bar{v}_k w_k (\mathbf{m}_k)_i)^2\right) \,,
\end{multline}
where 
\begin{equation*}
    n_k = \sum_{i=1}^{n} (\mathbf{m}_k)_i\,, \quad w_k=1- \sqrt\frac{\nicefrac{\sigma^2}{\tau^2}}{n_k + \nicefrac{\sigma^2}{\tau^2}}\,.
\end{equation*}
We then replace $\mathbf{m}$ with $\hat{\mathbf{m}}$ from the Gumbel-Softmax approximation.

\paragraph{Affine motion likelihood.} 
For the affine motion model, the full covariance matrix $\Sigma$ prevents significant further simplification. Instead, we transform the log-likelihood (\cref{e:sup:log_exp_marginal}) to an equivalent form involving only $3 \times 3$ matrices, for which the required determinant and inverse can be calculated analytically. To that end, we introduce the following auxiliary variables:
\begin{equation*}
    G_k = \begin{bmatrix}\mathbf{x}_k&\mathbf{y}_k&\mathbf{1}_{n_k}\end{bmatrix}\,,\quad
    P_k = \begin{bmatrix}G_k&0\\0&G_k\end{bmatrix}\,, \quad
    \Sigma^{-1} = \Lambda = \begin{bmatrix}\alpha&\beta\\\gamma&\delta\end{bmatrix}\,.
\end{equation*}
Then $S_k$ is
\begin{equation*}
S_k = \nicefrac{1}{\sigma^2}P_k^\top P_k + \Lambda = \begin{bmatrix}\nicefrac{1}{\sigma^2}G_k^\top G_k + \alpha&\beta\\\gamma&\nicefrac{1}{\sigma^2} G_k^\top G_k+\delta\end{bmatrix}\,,
\end{equation*}
with
\begin{equation*}
G_k^\top G_k= \begin{pmatrix}
\mathbf{x}_k^\top\mathbf{x}_k & \mathbf{x}_k^\top\mathbf{y}_k & \mathbf{x}_k^\top\mathbf{1}_{n_k}\\
\mathbf{x}_k^\top\mathbf{y}_k & \mathbf{y}_k^\top\mathbf{y}_k & \mathbf{y}_k^\top\mathbf{1}_{n_k}\\
\mathbf{x}_k^\top\mathbf{1}_{n_k} & \mathbf{y}_k^\top\mathbf{1}_{n_k} & \mathbf{1}_{n_k}^\top\mathbf{1}_{n_k}
\end{pmatrix}\,.
\end{equation*}
Using this, the determinant is
\begin{equation*}
\det S_k = \det(\nicefrac{1}{\sigma^2} G_k^\top G_k + \alpha - \beta(\nicefrac{1}{\sigma^2}G_k^\top G_k + \delta)^{-1}\gamma)\det(\nicefrac{1}{\sigma^2}G_k^\top G_k + \delta).
\end{equation*}
Similarly, the inverse is then
\begin{align*}
S_k^{-1} &= \begin{bmatrix}A_k & B_k \\ C_k & D_k\end{bmatrix}\,, \quad \textrm{where} \\
D_k &= (\nicefrac{1}{\sigma^2}G_k^\top G_k+ \delta - \gamma(\nicefrac{1}{\sigma^2}G_k^\top G_k+ \alpha)^{-1}\beta)\\
C_k &= -D_k \gamma(\nicefrac{1}{\sigma^2}G_k^\top G_k+ \alpha)^{-1}\\
B_k &= -(\nicefrac{1}{\sigma^2}G_k^\top G_k+ \alpha)^{-1}\beta D_k\\
A_k &= (\nicefrac{1}{\sigma^2}G_k^\top G_k+ \alpha)^{-1} -B_k\gamma(\nicefrac{1}{\sigma^2}G_k^\top G_k+ \alpha)^{-1} \mathrm{,~such~that}\\
\mathbf{h}_k &= \begin{pmatrix}
\mathbf{u}_k^\top \mathbf{x}_k & \mathbf{u}_k^\top \mathbf{y}_k & \mathbf{u}_k^\top \mathbf{1}_{n_k}
\end{pmatrix}\\
\mathbf{r}_k &= \begin{pmatrix}
\mathbf{v}_k^\top \mathbf{x}_k & \mathbf{v}_k^\top \mathbf{y}_k & \mathbf{v}_k^\top \mathbf{1}_{n_k}
\end{pmatrix}\\
\mathbf{F}_k^\top P_k S_k^{-1} P_k^\top\mathbf{F}_k &= 
\mathbf{h}_kA_k\mathbf{r}_k^\top + 
\mathbf{r}_kC_k\mathbf{h}_k^\top +
\mathbf{h}_kB_k\mathbf{r}_k^\top +
\mathbf{r}_kD_k\mathbf{r}_k^\top.
\end{align*}
We implement inner products under Gumbel-Softmax as $\mathbf{a}_k^\top\mathbf{b}_k = \sum_{i=1}^n (\mathbf{a})_i(\mathbf{b})_i(\hat{\mathbf{m}}_k)_i$, for some vectors $\mathbf{a},\mathbf{b}$. The coordinate vectors $\mathbf{x}_k,\mathbf{y}_k$ have the origin set to the centroid of the predicted region $\begin{psmallmatrix}x_{c,k} \\ y_{c,k}\end{psmallmatrix} = n_k^{-1}\begin{psmallmatrix} \mathbf{x}^\top\mathbf{m}_k \\ \mathbf{y}^\top\mathbf{m}_k\end{psmallmatrix}$. The expressions can then be substituted back to \cref{e:sup:log_exp_marginal}. We show implementation in \cref{alg:sup_neg_log_like}.

\begin{algorithm}
\caption{Implementation of negative flow likelihood $ -\log p( \mathbf{f} \mid \mathbf{m})$ under affine motion prior. Key quantities in the inner loop are underlined.}\label{alg:sup_neg_log_like}
\begin{algorithmic}[1]
\Procedure{NLL}{mean $\mu$, covariance $\Sigma$, variance $\sigma^2$, flow $\mathbf{f}$, masks $\mathbf{m}_k$, height $H$, width $W$}
\State $(\mu_1~\mu_2~\mu_3~\mu_4~\mu_5~\mu_6) \gets \mu$
\State $\mathbf{x},\mathbf{y} \gets \mathrm{lattice}(H, W)$
\State $\begin{bsmallmatrix}\mathbf{u} \\ \mathbf{v} \end{bsmallmatrix} \gets \mathbf{f}$
\State $\Lambda \gets \Sigma^{-1}$
\State $\begin{bsmallmatrix}\alpha&\beta\\\gamma&\delta\end{bsmallmatrix} \gets \Lambda$
\ForAll{$k$ regions}
    \State $n_k \gets \sum_i (\mathbf{m}_k)_i$
    \State $\mathbf{\hat x}_k \gets \mathbf{x} - n_k^{-1}\mathbf{x}^\top\mathbf{m}_k$ \Comment{Set origin to centroid}
    \State $\mathbf{\hat y}_k \gets \mathbf{y} - n_k^{-1}\mathbf{y}^\top\mathbf{m}_k$
    \State $\mathbf{x}_k^\top\mathbf{x}_k \gets \sum_i (\mathbf{\hat x}_k)_i^2(\mathbf{m}_k)_i$
    \State $\mathbf{x}_k^\top\mathbf{y}_k \gets \sum_i (\mathbf{\hat x}_k)_i(\mathbf{\hat y}_k)_i(\mathbf{m}_k)_i$
    \State $\mathbf{y}_k^\top\mathbf{y}_k \gets \sum_i (\mathbf{\hat y}_k)_i^2(\mathbf{m}_k)_i$
    \State $\mathbf{x}_k^\top\mathbf{1}_{n_k} \gets \sum_i (\mathbf{\hat x}_k)_i(\mathbf{m}_k)_i$
    \State $\mathbf{y}_k^\top\mathbf{1}_{n_k} \gets \sum_i (\mathbf{\hat y}_k)_i(\mathbf{m}_k)_i$
    \State $G_k^\top G_k \gets \begin{pmatrix}
\mathbf{x}_k^\top\mathbf{x}_k & \mathbf{x}_k^\top\mathbf{y}_k & \mathbf{x}_k^\top\mathbf{1}_{n_k}\\
\mathbf{x}_k^\top\mathbf{y}_k & \mathbf{y}_k^\top\mathbf{y}_k & \mathbf{y}_k^\top\mathbf{1}_{n_k}\\
\mathbf{x}_k^\top\mathbf{1}_{n_k} & \mathbf{y}_k^\top\mathbf{1}_{n_k} & n_k
\end{pmatrix}$
    \State $D_k \gets (\nicefrac{1}{\sigma^2}G_k^\top G_k+ \delta - \gamma(\nicefrac{1}{\sigma^2}G_k^\top G_k+ \alpha)^{-1}\beta)$
    \State $C_k \gets -D_k \gamma(\nicefrac{1}{\sigma^2}G_k^\top G_k+ \alpha)^{-1}$
    \State $B_k \gets -(\nicefrac{1}{\sigma^2}G_k^\top G_k+ \alpha)^{-1}\beta D_k$
    \State $A_k \gets (\nicefrac{1}{\sigma^2}G_k^\top G_k+ \alpha)^{-1} -B_k\gamma(\nicefrac{1}{\sigma^2}G_k^\top G_k+ \alpha)^{-1}$
    
    \State $\mathbf{u}_k^\top\mathbf{x}_k \gets \sum_i ((\mathbf{u})_i)((\mathbf{x})_i - x_{c,k})(\mathbf{m}_k)_i$
    
    \State $\mathbf{\hat u}_k \gets \mathbf{u} - ((\mu_1 - 1)\mathbf{\hat x} + \mu_2 \mathbf{\hat y} + \mu_3)$  \Comment{Center flow according to mean motion}
    \State $\mathbf{\hat v}_k \gets \mathbf{v} - ((\mu_5 - 1)\mathbf{\hat y} + \mu_4 \mathbf{\hat x} + \mu_6)$
    
    \State $\underline{\mathbf{F}_k^\top \mathbf{F}_k} \gets \sum_i (\mathbf{\hat u}_k)_i^2(\mathbf{m}_k)_i + \sum_i (\mathbf{\hat v}_k)_i^2(\mathbf{m}_k)_i$
    
    \State $\mathbf{u}_k^\top\mathbf{x}_k \gets \sum_i (\mathbf{\hat u}_k)_i(\mathbf{\hat x}_k)_i(\mathbf{m}_k)_i$
    \State $\mathbf{u}_k^\top\mathbf{y}_k \gets \sum_i (\mathbf{\hat u}_k)_i(\mathbf{\hat y}_k)_i(\mathbf{m}_k)_i$
    \State $\mathbf{u}_k^\top\mathbf{1}_{n_k} \gets \sum_i (\mathbf{\hat u}_k)_i(\mathbf{m}_k)_i$
    
    \State $\mathbf{h}_k \gets \begin{pmatrix}\mathbf{u}_k^\top \mathbf{x}_k & \mathbf{u}_k^\top \mathbf{y}_k & \mathbf{u}_k^\top \mathbf{1}_{n_k}
    \end{pmatrix}$
    
    \State $\mathbf{v}_k^\top\mathbf{x}_k \gets \sum_i (\mathbf{\hat v}_k)_i(\mathbf{\hat x}_k)_i(\mathbf{m}_k)_i$
    \State $\mathbf{v}_k^\top\mathbf{y}_k \gets \sum_i (\mathbf{\hat v}_k)_i(\mathbf{\hat y}_k)_i(\mathbf{m}_k)_i$
    \State $\mathbf{v}_k^\top\mathbf{1}_{n_k} \gets \sum_i (\mathbf{\hat v}_k)_i(\mathbf{m}_k)_i$
    
    \State $\mathbf{r}_k \gets \begin{pmatrix}
    \mathbf{v}_k^\top \mathbf{x}_k & \mathbf{v}_k^\top \mathbf{y}_k & \mathbf{v}_k^\top \mathbf{1}_{n_k}
    \end{pmatrix}$
     
    \State $\underline{\mathbf{F}_k^\top P_k S_k^{-1} P_k^\top\mathbf{F}_k} \gets 
    \mathbf{h}_kA_k\mathbf{r}_k^\top + 
    \mathbf{r}_kC_k\mathbf{h}_k^\top +
    \mathbf{h}_kB_k\mathbf{r}_k^\top +
    \mathbf{r}_kD_k\mathbf{r}_k^\top$
    
    \State $\underline{\det S_k} \gets \det(\nicefrac{1}{\sigma^2} G_k^\top G_k + \alpha - \beta(\nicefrac{1}{\sigma^2}G_k^\top G_k + \delta)^{-1}\gamma)\det(\nicefrac{1}{\sigma^2}G_k^\top G_k + \delta)$
\EndFor
\State $\mathbf{return~~} HW\log(2\pi\sigma^2) +\frac{1}{2}\sum_k\log\frac{\det S_k}{\det \Lambda} + \frac{1}{2\sigma^2}\sum_k (\mathbf{F}_k^\top\mathbf{F}_k - \mathbf{F}_k^\top P_k S_k^{-1} P_k^\top\mathbf{F}_k)$
\EndProcedure
\end{algorithmic}
\end{algorithm}

\subsection{Further justification}

We consider whether the inclusion of the prior on the motion parameters offers any benefits. Consider a simple translation-only model. Since objects are only translating, each pixel in a region should be very close to the mean translation of that region. We can assess the mean for a region as $\begin{bsmallmatrix}\bar{u}_k\mathbf{m}_k\\ \bar{v}_k\mathbf{m}_k\end{bsmallmatrix}$, considering some variance $\sigma^2$ around it:
\begin{align*}
    \log\hat{p}(\mathbf{f}\mid\mathbf{m}) &= \log\mathcal{N}(\begin{bsmallmatrix}
        \mathbf{u}\\
        \mathbf{v}
    \end{bsmallmatrix};\sum_k\begin{bsmallmatrix}
        \bar{u}_k\mathbf{m}_k\\
        \bar{v}_k\mathbf{m}_k
    \end{bsmallmatrix}, \sigma^2I)\\
&=
-n\log(2\pi\sigma^2) 
-\frac{1}{2\sigma^2}\sum_{i=1}^{n}\left((u_i - \sum_k \bar{u}_k(\mathbf{m}_k)_i)^2 + (v_i - \sum_k \bar{v}_k(\mathbf{m}_k)_i)^2\right).
\end{align*}
Such model, up to a scaling factor, was already considered for features~\citep{choudhury2021unsupervised} and optical flow~\citep{choudhury+karazija2022guess}.
This expression for $\log\hat{p}(\mathbf{f}\mid\mathbf{m})$ can be compared with our version of translation only model \cref{e:sup:trans_loglike}.
By considering the prior on the motion parameters, we introduce a weighing factor $w_k$ for each mean $\bar{u}_k\mathbf{m}_k$, which discounts the contribution of smaller segments. Similarly, the term $\sum_k\log\frac{n_k +\nicefrac{\sigma^2}{\tau^2}}{\nicefrac{\sigma^2}{\tau^2}} \approx \sum_k\log{n_k}$ encourages larger masks, since $\sum_k n_k = n$. The prior helps to encode that larger regions should be preferred.

\section{\textsc{MovingClevrTex} and \textsc{MovingClevr}}

We extend the implementation of~\cite{karazija2021clevrtex} to generate video datasets of \textsc{CLEVR} and \textsc{ClevrTex} scenes. We follow original sampling set up of \textsc{ClevrTex} -- each scene contains a random arrangement of 3--10 objects. We uniformly choose between scenarios where a single random objects is given initial motion, two random objects are provided initial motion or all objects are moving. We sample a random initial translation in XY plane for the object between keyframes 0 and 3, which builds momentum. Physics simulation takes over from keyframe 4. Mass of each object is set to equal its scale (numerically), and we use value of 0.1 for the `bounciness' parameters. This results in objects sliding, rotating due to shape and friction, and colliding. Collisions can make other objects move. Each simulation is 5 frames long and we render keyframes 4 to 8.

We sample 10000 scenes for \textsc{MovingClevrTex}, which gives the same number of frames as the original \textsc{ClevrTex} (where each scene had only single frame). For \textsc{MovingClevr}, we sample 5000 scenes. 1000 and 500 scenes are kept as validation for \textsc{MovingClevrTex} and for \textsc{MovingClevr}, respectively. We use the same rendering and lighting parameters as in ~\cite{karazija2021clevrtex}, except we slightly reduce the scale of the surface displacement mapping for the background. This reduces the visibility of clipping of the detailed object geometry with the background geometry, which might occur due to physics simulation working on simplified meshes.

\section{Hyperparameters}\label{s:sup_hparams}

Our method can use any segmentation network $\Phi$. We employ the recent Mask2Former\footnote{Code available from \url{https://github.com/facebookresearch/Mask2Former}.} architecture, using 6-layer CNN backbone from \citep{locatello2020object,kipf2022conditional} for simulated datasets and ResNet-18 for KITTI in the main experiments. We also experiment with Swin-tiny transformer as the backbone, as it offers balanced performance in both visually simple and complex settings. We ablate these choices in \cref{s:sup_ablations}.

The networks are trained with AdamW~\cite{loshchilov2017decoupled}, with a learning rate of $3 \times 10^{-6}$ and batch size of 32, for 250k iterations.
We employ gradient clipping when the 2-norm exceeds 0.01 and linear learning rate warm-up for the first 5k iterations to stabilize the training.
The learning rate is reduced by a factor of 10 after 200k iterations.
When training with warping loss on MOVi datasets, we found it beneficial to train for longer, for 500k iterations, reducing the learning rate by factor of 10 also after 400k iterations.
We found it beneficial to linearly anneal $\beta$ from 0.1 to -0.1 over 5k iterations, encouraging the network to explore initially but focus on low-entropy distributions in the end. We found this had the effect of encouraging the network to assign background pixels to a single slot. 

For the prior, we set $\sigma^2$ (\cref{e:sup:log_exp_marginal}) to $0.5$ and use $\mu = (1~0~0~0~1~0)^\top$. We use the following covariance matrix
\begin{align*}
\Sigma &=
\begin{pmatrix}
 0.005 &    0 &  0 &    0 &     0 &  0 \\
     0 & 0.05 &  0 &    0 &     0 &  0 \\
     0 &    0 & 15 &    0 &     0 &  0 \\
     0 &    0 &  0 & 0.05 &     0 &  0 \\
     0 &    0 &  0 &    0 & 0.005 &  0 \\
     0 &    0 &  0 &    0 &     0 & 15 
\end{pmatrix}.
\end{align*}

\subsection{Settings in ablations}

 When experimenting with translation-only model, we use the same parameters and settings where possible. We set $\tau^2$, so $0.5, 16.0, 34.0$ for \textsc{CLEVR}/\textsc{ClevrTex}, \movia, and \movic, respectively (values picked from specialized covariance matrices, see \cref{s:sup_spec_cov}).

For GNM~\citep{jiang2020generative}, we use implementation and parameters described in \citep{karazija2021clevrtex}. Owning to our loss being lower bound on log-probability, we simply add our loss to existing ELBO loss with hyperparameters above, only changing $\sigma^2 = 0.1$. We hypothesize that lower noise model is beneficial, as it provides stronger learning signal in the early stages when reconstruction is noisy due to untrained VAEs. The overfitting to errors of the motion approximation is handled, instead, by the networks balancing between appearance reconstruction and motion explanation objectives.

For SA~\citep{locatello2020object}, we also use implementation of \citep{karazija2021clevrtex}. SA loss is multiplied by 100 before adding our formulation.

\subsection{Settings in KITTI}

For experiments on KITTI, we replace the backbone to ResNet-18 to match prior work. We reduce batch size to 8, increase learning rate to $10^{-4}$. Learning rate is linearly warmed up for 10 iterations, and reduced by a factor of 10 after 5500 iterations. We employ backbone learning rate multiplier of 0.1. All other settings as before.

\subsection{Model parameters of comparisons}
Here we describe implementation and hyperparameters used for comparative methods in our experiments.

\paragraph{SAVi~\cite{kipf2022conditional}.}
We follow the parameters given for \textit{conditional} SAVi-S in their code repository\footnote{Code available from \url{https://github.com/google-research/slot-attention-video}.}, except to make SAVi unconditional we unset the conditioning key and make the slots to be learnable parameters.
\paragraph{SCALOR~\cite{jiang2020scalor}.}
We use the optimized SCALOR parameters mentioned by SAVi~\cite{kipf2022conditional} to train SCALOR. Particularly we use the MOVI dataset parameters to train \movia and MOVI++ parameters to train \movic.

\paragraph{GWM~\cite{choudhury+karazija2022guess}.} For GWM, we follow the parameters mentioned in the paper, except we do not employ spectral clustering and match the number of components to our settings for each dataset.

\section{Additional ablations}\label{s:sup_ablations}
\begin{table}[t]
\centering
\caption{Supplementary ablations for our methods. We show the impact segmentation networks have for different datasets \cref{tab:s:abl-network}. We further study the impact of our tuned covariance matrix in  \cref{tab:s:abl-cov}. Results with post-processing applied.}
\begin{subtable}{0.9\linewidth}
\caption{Choice of Model Architecture}
\label{tab:s:abl-network}
\vspace{-0.8em}
\resizebox{\textwidth}{!}{
\begin{tabular}{lccccccc} \\
\toprule
 & & \multicolumn{2}{c}{MOVi-A} & \multicolumn{2}{c}{MOVi-C} & \multicolumn{2}{c}{\textsc{MovingClevrTex}} \\
\cmidrule(l{2pt}r{2pt}){3-4}\cmidrule(l{2pt}r{2pt}){5-6}\cmidrule(l{2pt}r{2pt}){7-8}
\textbf{Architecture}& \textbf{\# Params}& \textbf{FG-ARI}$\uparrow$ & \textbf{mIoU}$\uparrow$  & \textbf{FG-ARI}$\uparrow$ & \textbf{mIoU}$\uparrow$ & \textbf{FG-ARI}$\uparrow$ & \textbf{mIoU}$\uparrow$ \\
\midrule
M2F (Swin-tiny) & 47M &  $83.48$ & $72.61$ & $58.59$ & $35.67$ & $88.80$ & $69.62$  \\
M2F (ResNet50) & 44M &  $83.44$ & $68.06$ & $60.32$ & $34.80$ & $90.40$ & $67.07$ \\
M2F (ResNet18) & 31M & $84.04$ & $67.48$ & $60.84$ & $35.69$ & $90.31$ & $67.33$ \\
MF (Swin-tiny) & 44M & $81.78$ & $71.28$ & $54.45$ & $33.67$ & $71.07$ & $51.06$ \\
U-Net & 31M &  $90.79$ & $82.85$ & $60.28$ & $26.62$ & $87.03$ & $39.66$ \\

\bottomrule
\end{tabular}
}
\end{subtable}

\begin{subtable}{0.55\linewidth}
\caption{Choice of Covariance Marix}
\label{tab:s:abl-cov}
\vspace{-0.8em}
\resizebox{\textwidth}{!}{
\begin{tabular}{lcccccc} \\
\toprule
 & \multicolumn{2}{c}{MOVi-A} & \multicolumn{2}{c}{MOVi-C} & \\
\cmidrule(l{2pt}r{2pt}){2-3}
\cmidrule(l{2pt}r{2pt}){4-5}
$\mathbf{\Sigma}$ & \textbf{FG-ARI}$\uparrow$ & \textbf{mIoU}$\uparrow$  & \textbf{FG-ARI}$\uparrow$ & \textbf{mIoU}$\uparrow$ \\
\midrule
Generic & $82.32$ & $71.70$ & $58.12$ & $35.79$ \\
Tuned &  $83.48$ & $72.61$ & $58.59$ & $35.67$ \\
\bottomrule
\end{tabular}
}
\end{subtable}

\end{table}
\paragraph{Segmentation network.} We study the effect of segmentation network in \cref{tab:s:abl-network}. 
We find that using a much simpler U-Net~\cite{ronneberger2015u} architecture is beneficial on \movia which contains visually plain scenes. 
The U-Net architecture, however, results in performance degradation on visually complex data.

We also consider a version of Mask2Former architecture that uses much deeper backbone, using Resnet-50 and Resnet-18, instead. We find that the deeper backbones leads to similar performance, indicating that our formulation is not architecture-specific. Finally, we change to MaskFormer architecture, matching the network architecture used in ~\cite{choudhury+karazija2022guess}, and use smaller Swin-tiny backbone. We find that our loss formulation leads to improved performance still.

\paragraph{Covariance matrix.}\label{s:sup_spec_cov} We investigate whether further improvements are possible using specialised versions of the prior. To that end, we offset the mean translation prior to account for the dominantly downward motion of objects on \movia/\movic. We use 
$\mu_{\mathrm{\movia}} = (1~0~0~0~1~1.5)^\top$ and $\mu_{\mathrm{\movic}} = (1~0~0~0~1~1.8)^\top$, respectively.\footnote{In our experiments, Y axis is pointing down and X is pointing right.}
We use the following specialized covariance matrices:

\begin{align*}
\Sigma_{\mathrm{\movia}} &=
\begin{pmatrix}
   0.006 & -0.00004 &  0 & 0.00004 &    0.001 &  0 \\
-0.00004 &     0.04 &  0 &   -0.01 & -0.00008 &  0 \\
       0 &        0 & 16 &       0 &        0 &  0 \\
 0.00004 &    -0.01 &  0 &    0.04 &  0.00004 &  0 \\
   0.001 & -0.00008 &  0 & 0.00004 &    0.006 &  0 \\
       0 &        0 &  0 &       0 &        0 & 14 
\end{pmatrix} \\
\Sigma_{\mathrm{\movic}} &=
\begin{pmatrix}
    0.02 &   0.00002 &  0 & 0.000009 &     0.002 &  0 \\
 0.00002 &      0.03 &  0 &   -0.009 & -0.000006 &  0 \\
       0 &         0 & 36 &        0 &         0 &  0 \\
0.000009 &    -0.009 &  0 &     0.04 &  -0.00007 &  0 \\
   0.002 & -0.000006 &  0 & -0.00007 &      0.02 &  0 \\
       0 &         0 &  0 &        0 &         0 & 34 
\end{pmatrix}\\
\end{align*}

We obtain the dataset specific covariance matrices by forming initial estimates using a method described below using only optical flow. 
We then overwrite the entries to encode our belief that translation should be independent. Finally, we further tuned the values through by taking one search step for \movia and \movic each. In our experiments, we found that increasing diagonal elements (variances) and decreasing off-diagonal elements produced slightly better results. 

\paragraph{Initial covariance estimation.} We form initial estimate for the dataset-specific covariance matrices used in ablation only to start hyperparameter search in a sensible range. This method relies on observation that to form an estimate (1) all objects from a frame are not required -- only some are sufficient. Furthermore, for the selected object candidates, (2) precise boundaries are not necessary. The method is as follows:
\begin{enumerate}
    \item We extract discontinuities from the flow using Sobel filtering and treat these as flow edges.
    \item We then only consider regions where the optical flow is larger than zero, identifying foreground.
    \item We subtract edge pixels from the candidate foreground mask. This attempts to disconnect any overlapping objects using discontinuity in optical flow. 
    \item We run connected components algorithm to identify candidate object regions.
    \item Within each region, using the motion model (Eq. (2)) we estimate $\hat\theta$ by forming least-squares solution. We only considered estimates from regions larger than 100 pixels (for numerical stability) and where the residual error was within the 90\% percentile.
    \item Initial covariance estimate $\hat\Sigma$ is formed by calculating sample covariance over the combined set of inliers and extra $n$ no-motion values to account for stationary regions.
\end{enumerate}
We apply this method on a subset of the data. $n$ is the size of the subset.

We find using the specialized settings above give slight improvement on most metrics (\cref{tab:s:abl-cov}), indicating that using more appropriate prior for the data further improves results.

\section{Additional results}\label{s:sup_qual_res}

\begin{table}
\caption{Expanded benchmark results on \textsc{CLEVR}, \textsc{ClevrTex}, \textsc{CAMO}, and \textsc{OOD} comparing FG-ARI and mIoU metrics. Results are a mean of 3 seed $(\pm \sigma)$. Methods above the line are trained on single images, while methods below train on videos.\gray{$^\dagger$ -- indicates post-processing.}}
\label{tab:s:clevrtex_expanded}
\resizebox{\textwidth}{!}{
\begin{tabular}{lrrrrrrrr}
\toprule
 & \multicolumn{2}{c}{\textsc{CLEVR}}
      & \multicolumn{2}{c}{\textsc{ClevrTex}} 
      & \multicolumn{2}{c}{\textsc{OOD}}
      & \multicolumn{2}{c}{\textsc{CAMO}}
\\
\cmidrule(l{2pt}r{2pt}){2-3}
\cmidrule(l{2pt}r{2pt}){4-5}
\cmidrule(l{2pt}r{2pt}){6-7}
\cmidrule(l{2pt}){8-9}

\textbf{Model}
& \multicolumn{1}{c}{\textbf{FG-ARI}\(\uparrow\)}
& \multicolumn{1}{c}{\textbf{mIoU}\(\uparrow\)}
& \multicolumn{1}{c}{\textbf{FG-ARI}\(\uparrow\)}
& \multicolumn{1}{c}{\textbf{mIoU}\(\uparrow\)}
& \multicolumn{1}{c}{\textbf{FG-ARI}\(\uparrow\)}
& \multicolumn{1}{c}{\textbf{mIoU}\(\uparrow\)}
& \multicolumn{1}{c}{\textbf{FG-ARI}\(\uparrow\)}
& \multicolumn{1}{c}{\textbf{mIoU}\(\uparrow\)}
\\
\midrule

SPAIR \citep{crawford2019spatially}
& \arif{77.13}{1.92} & \mioud{65.95}{4.02}
& \arif{0.00}{0.00} & \mious{0.00}{0.00}
& \arif{0.00}{0.00} & \mious{0.00}{0.00}
& \arif{0.00}{0.00} & \mious{0.00}{0.00} 
\\
SPAIR$^{\dagger}$
& \arif{77.05}{1.96} & \mioud{66.87}{9.65}
& \arif{0.00}{0.00} & \mious{0.00}{0.00}
& \arif{0.00}{0.00} & \mious{0.00}{0.00}
& \arif{0.00}{0.00} & \mious{0.00}{0.00}
\\

SPACE \citep{lin2020space}
& \arif{22.75}{14.04} & \mioud{26.31}{12.93}
& \arif{17.53}{4.13} & \mious{9.14}{3.46}
& \arif{12.71}{3.44} & \mious{6.87}{3.32}
& \arif{10.55}{2.09} & \mious{8.67}{3.50}
\\
SPACE$^{\dagger}$ 
& \arif{22.74}{14.03} & \mioud{27.00}{13.69}
& \arif{17.52}{4.12} & \mious{9.68}{4.10}
& \arif{12.71}{3.44} & \mious{7.20}{3.75}
& \arif{10.54}{2.08} & \mious{9.25}{3.95}
\\

GenV2 \citep{engelcke2021genesis}
& \arif{57.90}{20.38} & \mioud{9.48}{0.55}
& \arif{31.19}{12.41} & \mious{7.93}{1.53}
& \arif{29.04}{11.23} & \mious{8.74}{1.64}
& \arif{29.60}{12.84} & \mious{7.49}{1.67}
\\
GenV2$^{\dagger}$
& \arif{57.78}{21.12} & \mioud{10.76}{1.39}
& \arif{30.55}{14.27} & \mious{9.04}{0.63}
& \arif{28.41}{13.20} & \mious{9.96}{0.70}
& \arif{29.19}{14.55} & \mious{8.40}{1.00}
\\

MN \citep{smirnov2021marionette}
& \arif{72.12}{0.64} & \mious{56.81}{0.40}
& \arif{38.31}{0.70} & \mious{10.46}{0.10}
& \arif{37.29}{1.04} & \mious{12.13}{0.19} 
& \arif{31.52}{0.87} & \mious{8.79}{0.15}
\\
MN$^{\dagger}$
& \arif{72.08}{0.62} & \mious{57.61}{0.40}
& \arif{38.34}{0.73} & \mious{10.34}{0.12}
& \arif{37.28}{1.07} & \mious{11.97}{0.21}
& \arif{31.54}{0.87} & \mious{8.77}{0.18}
\\

MONet \citep{burgess2019monet} 
& \arif{54.47}{11.41} & \mioud{30.66}{14.87}
& \arif{36.66}{0.87} & \mious{19.78}{1.02}
& \arif{32.97}{1.00} & \mious{19.30}{0.37}
& \arif{12.44}{0.73} & \mious{10.52}{0.38}
\\
MONet$^{\dagger}$
& \arif{61.36}{7.33} & \mioud{45.61}{4.80}
& \arif{35.64}{1.17} & \mious{23.59}{0.29}
& \arif{31.51}{1.46} & \mious{23.04}{0.52}
& \arif{9.94}{0.50}  & \mious{11.31}{0.30}
\\

SA \citep{locatello2020object}
& \arif{95.89}{2.37} & \mioud{36.61}{24.83}
& \arif{62.40}{2.23} & \mious{22.58}{2.07}
& \arif{58.45}{1.87} & \mious{20.98}{1.59}
& \arif{57.54}{1.01} & \mious{19.83}{1.41}
\\
SA$^{\dagger}$
& \arif{94.88}{1.67} & \mioud{37.68}{26.56}
& \arif{61.60}{2.29} & \mious{21.96}{1.79}
& \arif{57.41}{1.92} & \mious{20.60}{1.45}
& \arif{56.85}{1.12} & \mious{19.42}{1.42}
\\

IODINE \citep{greff2019iodine}
& \arif{93.81}{0.76} & \mioud{45.14}{17.85}
& \arif{59.52}{2.20} & \mious{29.17}{0.75}
& \arif{53.20}{2.55} & \mious{26.28}{0.85}
& \arif{36.31}{2.57} & \mious{17.52}{0.75}
\\
IODINE$^{\dagger}$
& \arif{93.68}{0.83} & \mioud{44.20}{18.67}
& \arif{60.63}{2.50} & \mious{29.40}{1.10}
& \arif{54.92}{2.24} & \mious{27.96}{0.81}
& \arif{38.29}{1.40} & \mious{18.87}{0.52}
\\

eMORL \citep{emami2021efficient} 
& \arif{93.25}{3.24} & \mioud{50.19}{22.56}
& \arif{55.62}{2.12} & \mious{30.17}{2.60}
& \arif{49.21}{2.69} & \mious{25.03}{1.99}
& \arif{37.66}{8.41} & \mious{19.13}{4.88}
\\
eMORL$^{\dagger}$
& \arif{93.09}{2.68} & \mioud{49.28}{24.28}
& \arif{58.59}{1.96} & \mious{31.64}{2.22}
& \arif{51.97}{2.44} & \mious{26.91}{1.69}
& \arif{43.83}{7.34} & \mious{22.40}{4.35}
\\

DTI-S \citep{monnier2021dtisprites}
& \arif{89.54}{1.44} & \mioud{48.74}{2.17}
& \arif{79.90}{1.37} & \mious{33.79}{1.30}
& \arif{73.67}{0.98} & \mious{32.55}{1.08}
& \arif{72.90}{1.89} & \mious{27.54}{1.55}
\\
DTI-S$^{\dagger}$
& \arif{89.86}{1.78} & \mioud{53.38}{3.51}
& \arif{79.86}{1.36} & \mious{32.20}{1.49}
& \arif{73.60}{0.97} & \mious{30.74}{1.22}
& \arif{72.89}{1.88} & \mious{26.30}{1.57}
\\

GNM \citep{jiang2020generative}
& \arif{65.05}{4.19} & \mioud{59.92}{3.72}
& \arif{53.37}{0.67} & \mious{42.25}{0.18}
& \arif{48.43}{0.86} & \mious{40.84}{0.30} 
& \arif{15.73}{0.89} & \mious{17.56}{0.74}
\\
GNM$^{\dagger}$ 
& \arif{65.67}{4.23} & \mioud{63.38}{3.76}
& \arif{53.38}{0.67} & \mious{44.30}{0.19}
& \arif{48.44}{0.86} & \mious{42.87}{0.28}
& \arif{15.72}{0.89} & \mious{18.53}{0.75}
\\

\midrule

SAVi~\citep{kipf2022conditional} 
& \multicolumn{1}{c}{---} & \multicolumn{1}{c}{---}
& \arifnoerr{49.54} & \miousnoerr{31.88} 
& \arifnoerr{42.68} & \miousnoerr{30.31} 
& \arifnoerr{42.67} & \miousnoerr{29.60} 
\\
Ours
& \arif{91.69}{0.30} & \mioud{66.70}{0.32}
& \arif{90.80}{0.22} & \mious{55.07}{0.44}
& \arif{76.01}{0.56} & \mious{46.84}{0.20}
& \arif{72.78}{1.31} & \mious{42.30}{1.09}
\\

Ours $^{\dagger}$
& \arifbf{95.94}{0.43} & \mioudbf{84.86}{4.06}
& \arifbf{92.61}{0.22} & \miousbf{77.67}{0.25}
& \arifbf{78.24}{0.43} & \miousbf{55.54}{0.44}
& \arifbf{77.43}{0.86} & \miousbf{56.43}{0.80}
\\

\bottomrule
\end{tabular}
}
\end{table}
\paragraph{Expanded results on \textsc{Clevr} and \textsc{ClevrTex}.} In \cref{tab:s:clevrtex_expanded} we show expanded version of the results on \textsc{Clevr} and \textsc{ClevrTex} benchmarks.

\begin{figure}[t]
    \centering
    \includegraphics[page=4,width=\linewidth,trim={0 4.0cm 0 0},clip]{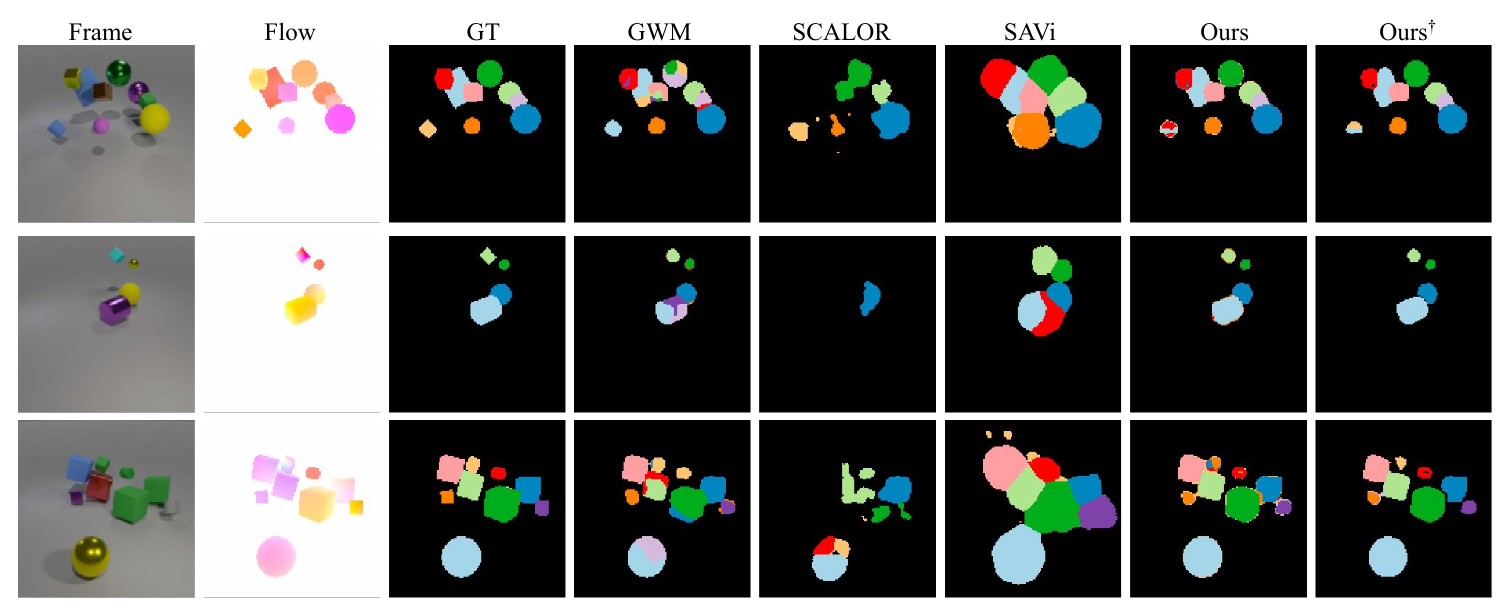} 
    \includegraphics[page=5,width=\linewidth,trim={0 4.7cm 0 0.8cm},clip]{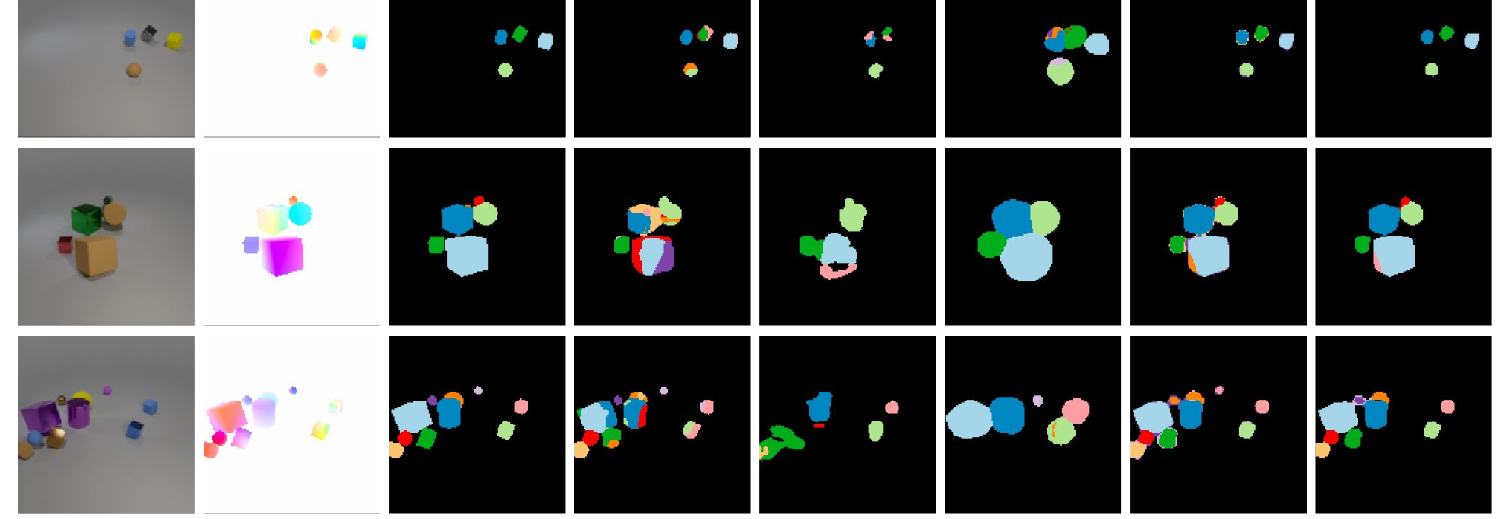} 
    \includegraphics[page=6,width=\linewidth,trim={0 4.5cm 0 0.8cm},clip]{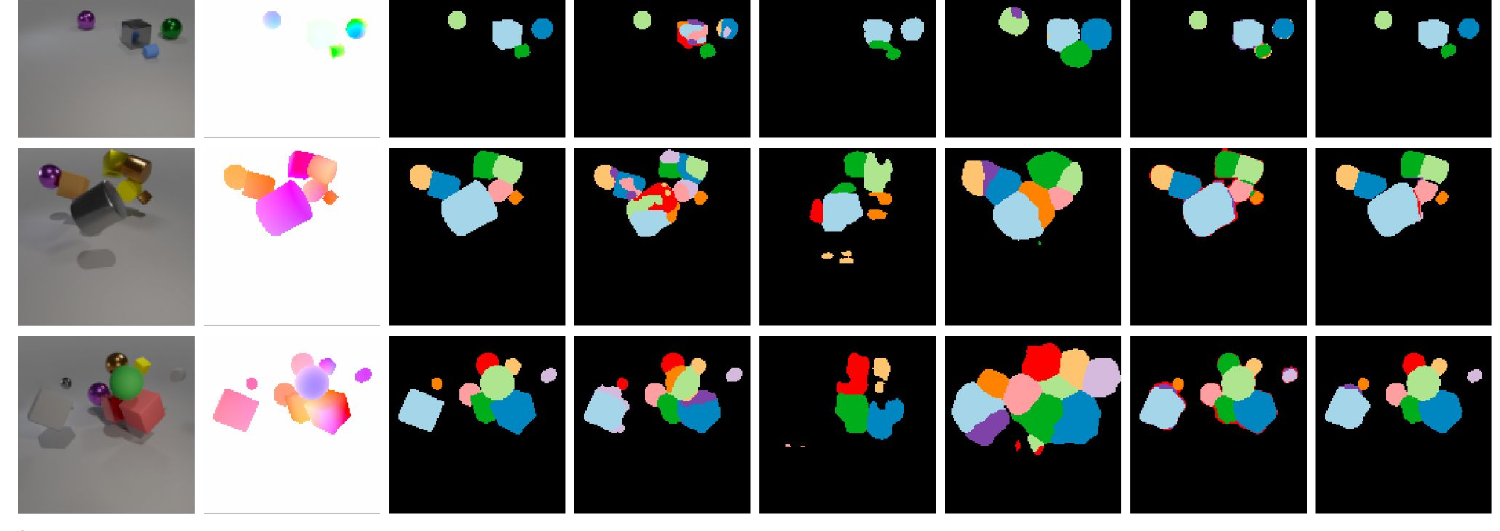} 
    \includegraphics[page=7,width=\linewidth,trim={0 4.5cm 0 0.8cm},clip]{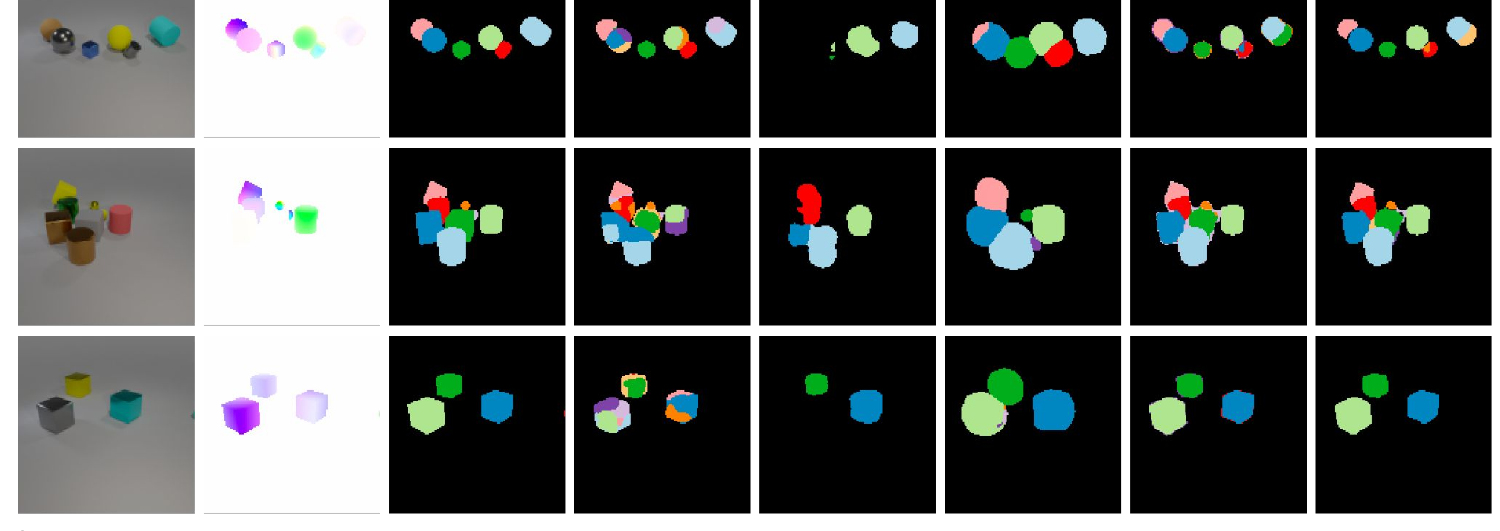} 
    \caption{Additional qualitative comparison on \movia.
     Our method performs consistently well compared to other methods. 
     \gray{${}^\dagger$-- indicates post-processing.}}
    \label{fig:supp_qualitative_video_movi_a}
\end{figure}

\begin{figure}[t]
    \centering    \includegraphics[page=8,width=\linewidth,trim={0 4.3cm 0 0},clip]{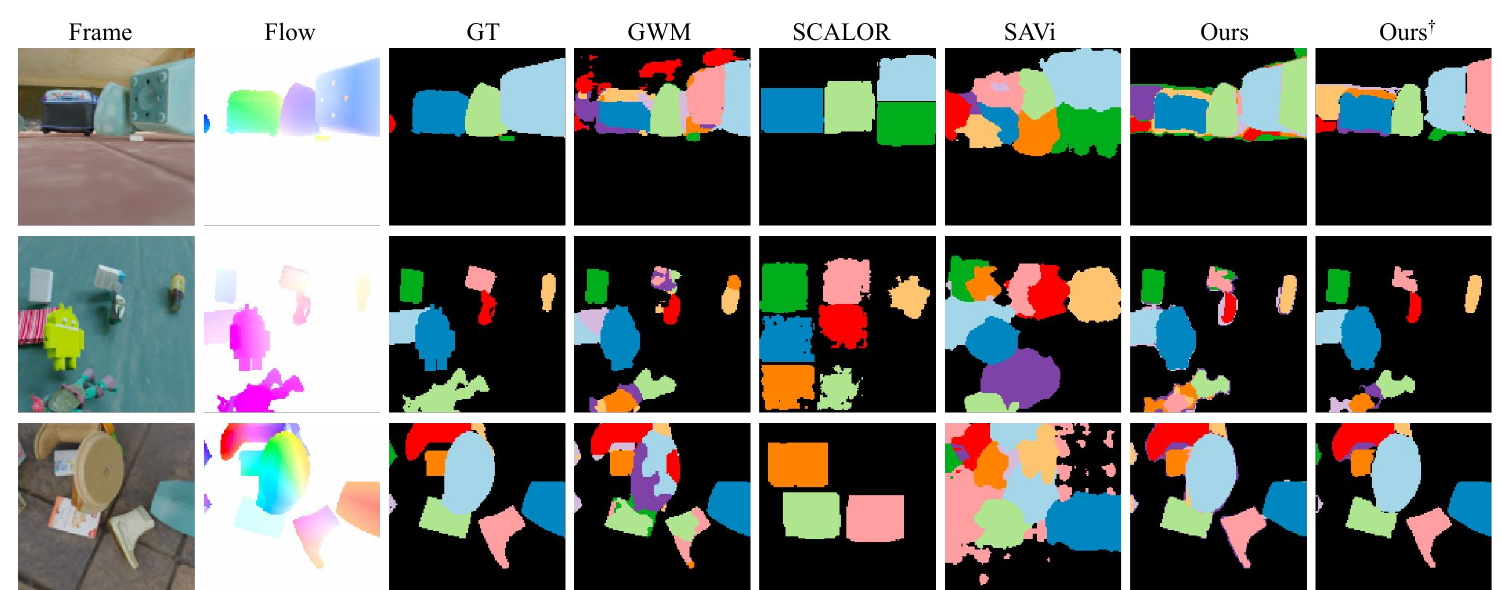} 
    \includegraphics[page=9,width=\linewidth,trim={0 4.7cm 0 0.8cm},clip]{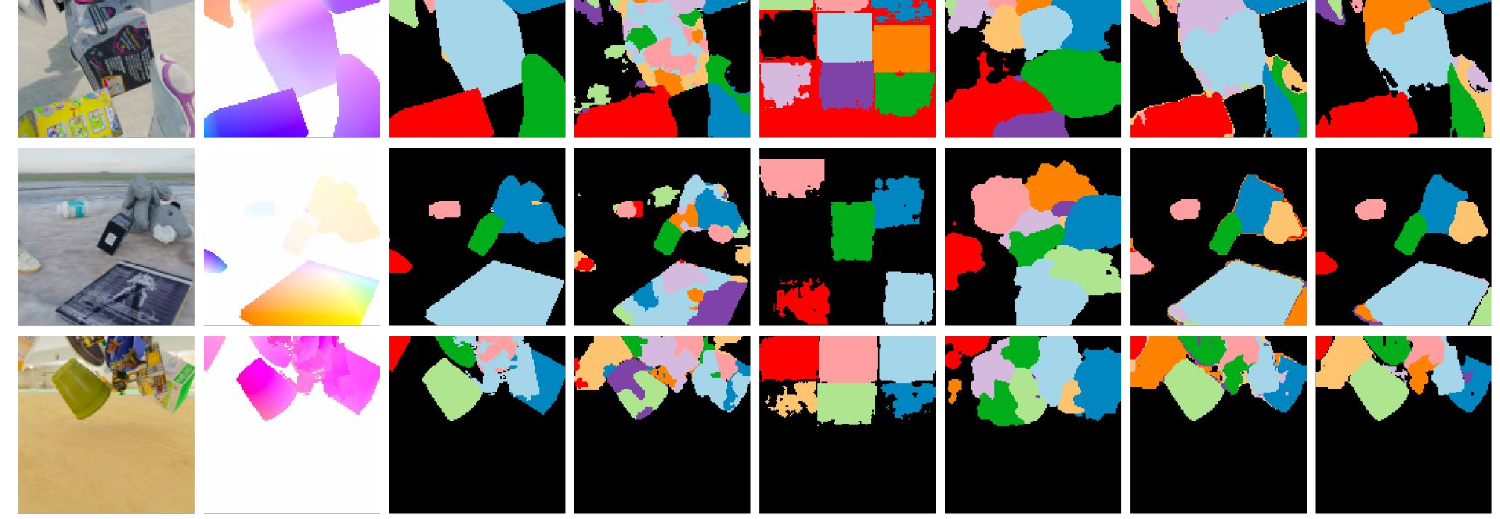} 
    \includegraphics[page=10,width=\linewidth,trim={0 4.7cm 0 0.8cm},clip]{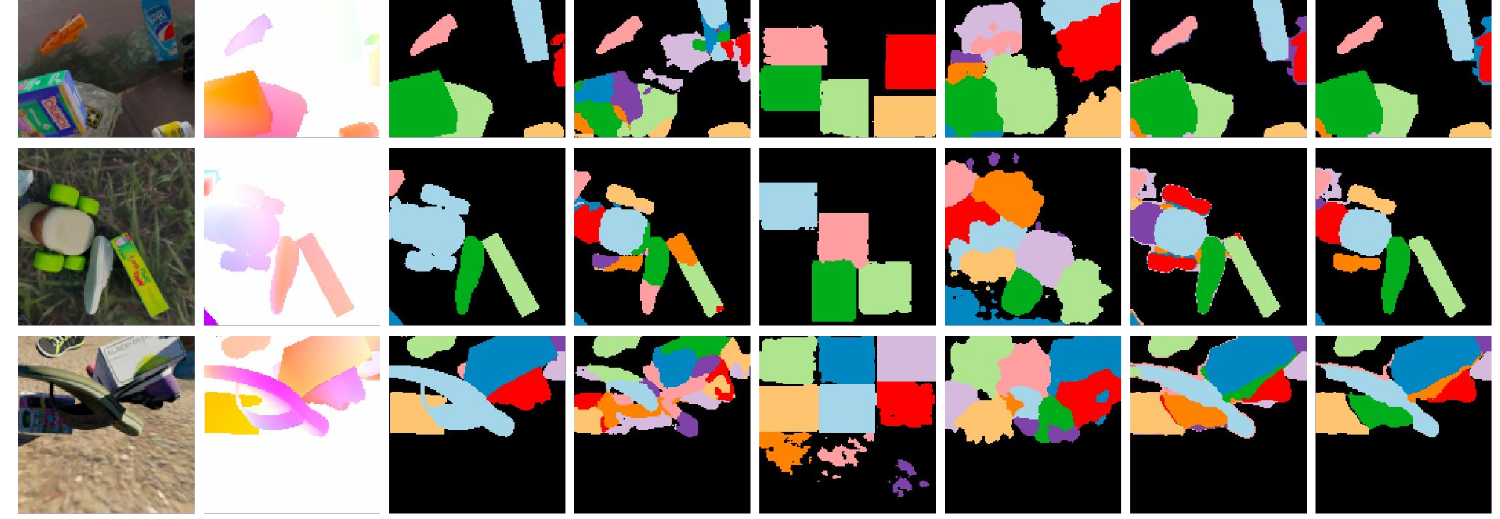} 
    \includegraphics[page=11,width=\linewidth,trim={0 4.5cm 0 0.8cm},clip]{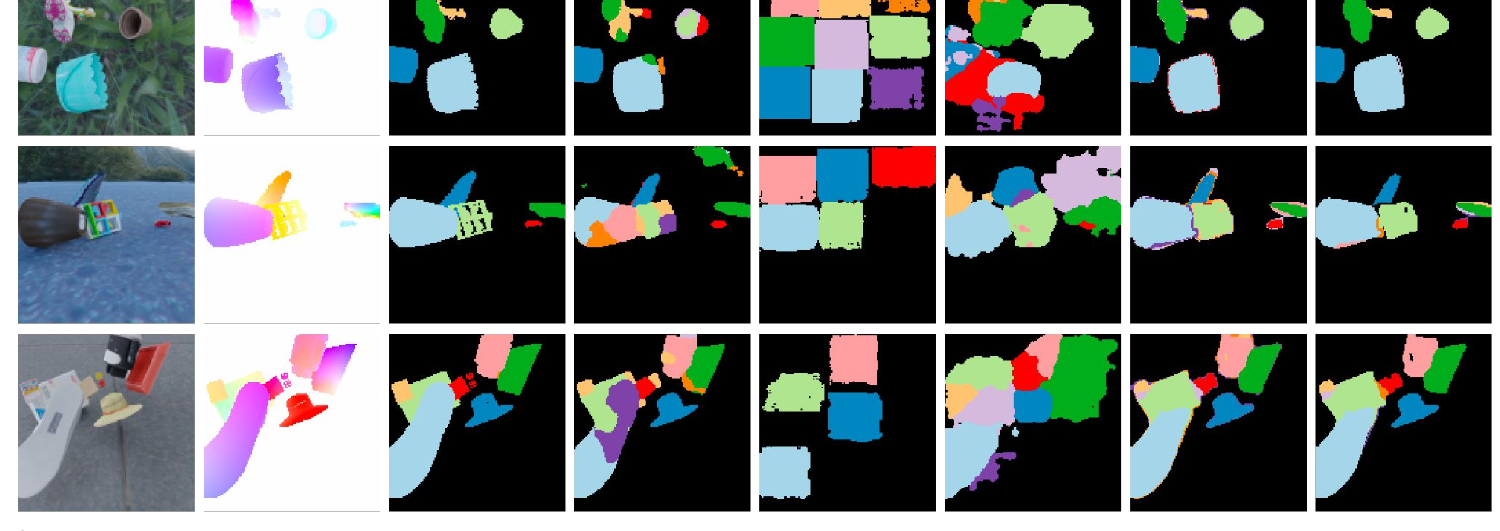} 
    \caption{Additional qualitative comparison on \movic.
     \gray{${}^\dagger$-- indicates post-processing.} }
    \label{fig:supp_qualitative_video_movi_c}
\end{figure}

\begin{figure}[t]
    \centering
    \includegraphics[page=12,width=\linewidth,trim={0 4.2cm 0 0cm},clip]{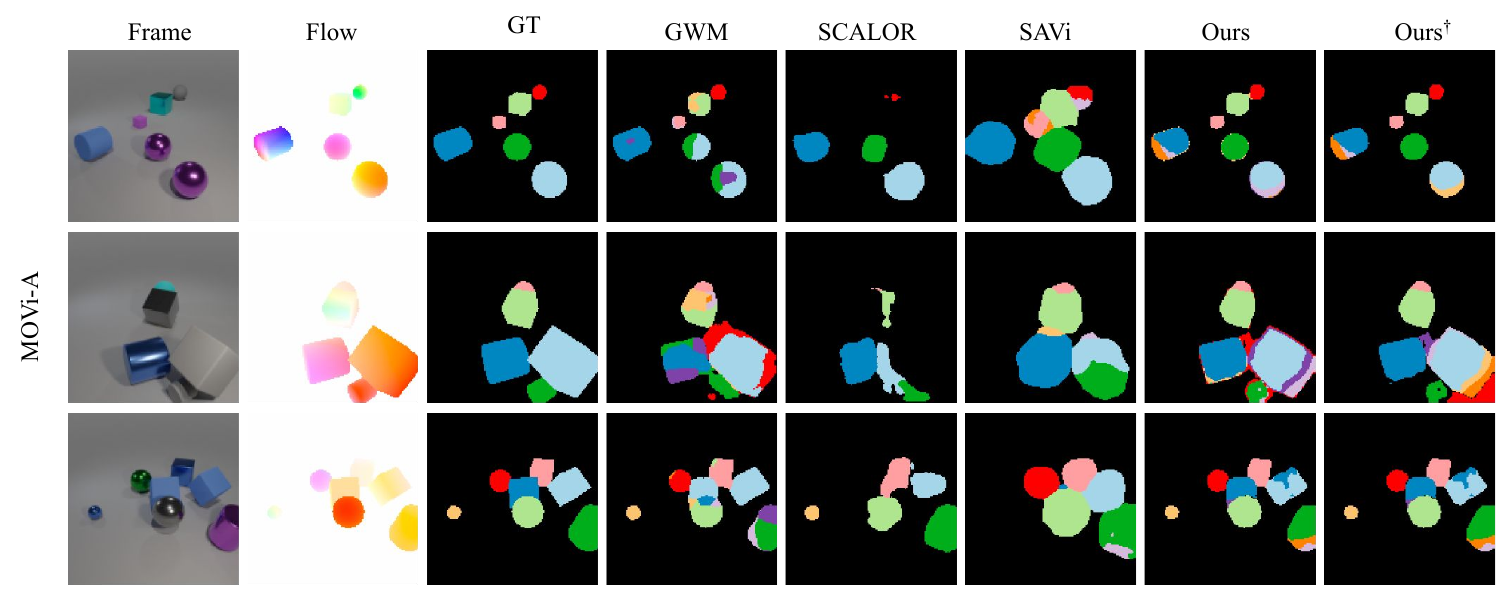}  \\[2pt]
    \includegraphics[page=13,width=\linewidth,trim={0 4.8cm 0 0.8cm},clip]{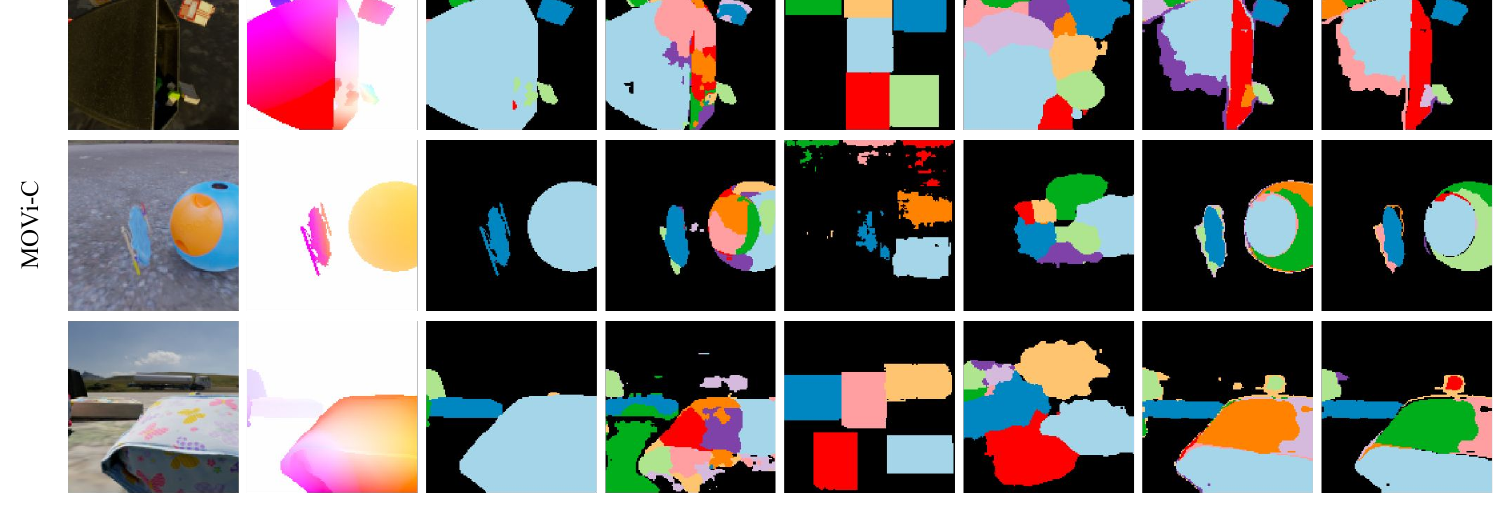} 
    \caption{Additional examples of failure cases on \movia and \movic. \gray{${}^\dagger$-- indicates post-processing.}}
    \label{fig:supp_qualitative_video_failure}
\end{figure}

\paragraph{Qualitative results on \movia and \movic.} In ~\cref{fig:supp_qualitative_video_movi_a} and ~\ref{fig:supp_qualitative_video_movi_c} we show additional qualitative results on \movia and \movic respectively. Following the results in main paper, the segments discovered by our method are semantically meaningful. Our object boundaries are of higher quality than the comparable methods. GWM suffers from oversegmentation of the objects. SCALOR has dificulty with complex datasets, such as \movic as observed in \cref{fig:supp_qualitative_video_movi_c}. SAVI's object boundaries do not conform to object shape. We also provide additional failure cases of our model in \cref{fig:supp_qualitative_video_failure}. Our model has difficulty with objects that have complex motion for our affine formulation to model ably.

\end{document}